\pdfoutput=1

\documentclass[11pt]{article}
\usepackage{CJKutf8}
\usepackage[preprint]{acl}
\usepackage{anyfontsize}
\usepackage{amsmath}
\usepackage{bbding}
\usepackage{booktabs}
\usepackage{times}
\usepackage{latexsym}
\usepackage{makecell}
\usepackage{hyperref}
\usepackage{algpseudocode}
\usepackage{xcolor}
\usepackage{amssymb}
\usepackage{pifont}
\usepackage[T1]{fontenc}

\usepackage[utf8]{inputenc}

\usepackage{microtype}

\usepackage{inconsolata}

\usepackage{graphicx}
\usepackage[ruled]{algorithm2e}
\title{CMMaTH: A Chinese Multi-modal Math Skill Evaluation Benchmark for Foundation Models}

\author{
    Zhong-Zhi Li\textsuperscript{1,2}\thanks{~~~Equal Contribution} , Ming-Liang Zhang\textsuperscript{1,2}\footnotemark[1], Fei Yin\textsuperscript{1,2}, \\
    \textbf{Zhi-Long Ji\textsuperscript{3}}, \textbf{Jin-Feng Bai\textsuperscript{3}}, \textbf{Zhen-Ru Pan\textsuperscript{3}}, \\
    \textbf{Fan-Hu Zeng\textsuperscript{1,2}}, \textbf{Jian Xu\textsuperscript{1,2}}, \textbf{Jia-Xin Zhang\textsuperscript{1,2}}, \textbf{Cheng-Lin Liu\textsuperscript{1,2}}\thanks{~~~Corresponding Author}\\
    School of Artifcial Intelligence, University of Chinese Academy of Sciences\textsuperscript{1} \\
    MAIS, Institute of Automation of Chinese Academy of Sciences\textsuperscript{2}, 
    Tomorrow Advancing Life\textsuperscript{3}\\
    \{lizhongzhi2022, zhangmingliang2018\}@ia.ac.cn, \\
    \{jizhilong, baijinfeng, panzhenru,\}@tal.com, \\
    \{fyin, liucl\}@nlpr.ia.ac.cn
}

\begin{document}
\begin{CJK}{UTF8}{gbsn}
\maketitle

\begin{abstract}
Due to the rapid advancements in multimodal large language models, evaluating their multimodal mathematical capabilities continues to receive wide attention. Despite the datasets like MathVista proposed benchmarks for assessing mathematical capabilities in multimodal scenarios, there is still a lack of corresponding evaluation tools and datasets for fine-grained assessment in the context of K12 education in Chinese language. To systematically evaluate the capability of multimodal large models in solving Chinese multimodal mathematical problems, we propose a Chinese Multi-modal Math Skill Evaluation Benchmark, named CMMaTH, contraining 23k multimodal K12 math related questions, forming the largest Chinese multimodal mathematical problem benchmark to date. CMMaTH questions from elementary to high school levels, provide increased diversity in problem types, solution objectives, visual elements, detailed knowledge points, and standard solution annotations. We have constructed an open-source tool GradeGPT  integrated with the CMMaTH dataset, facilitating stable, rapid, and cost-free model evaluation. Our data and code are available.
\end{abstract}

\section{Introduction}

Large language models(LLMs) excel in various language tasks, while multimodal models effectively handle visual-language problems. They advance natural language processing and computer vision fields, providing powerful solutions for complex tasks. Multimodal large models demonstrate potential as versatile solvers for multimodal problems. 

The systematic evaluation of large models' performance across various mathematical reasoning scenarios has been a subject of extensive research. GSM8K and MATH\cite{GSM8K, MATH} assessed the ability in multi-step mathematical reasoning by constructing a high-quality set of elementary school math word problems or various competition mathematics problems. By collecting a diverse set of mathematical problems containing both textual and visual components, \citet{MathVista, MATH-VISION, GeoEval} systematically evaluated the ability of large multimodal models to perceive visual elements and solve corresponding multimodal problems. \citet{MGSM} constructed a multilingual mathematical reasoning dataset, MGSM, for evaluating the LLM reasoning ability in multilingual environments. 

However, in non-English multimodal contexts, especially in Chinese scenarios, there is still a lack of sufficiently detailed and diverse benchmarks for assessing mathematical abilities. To assess the capability of large language models in non-English contexts, \citet{C-Eval} and \citet{CMMMU} constructed multidisciplinary Chinese question answering datasets C-Eval and CMMMU to evaluate the knowledge and reasoning abilities of multimodal large models. However, C-Eval lacks evaluation in multimodal contexts, while CMMMU's dataset has relatively low diversity, consisting of only 540 questions.

\begin{figure*}
  \begin{center}
  \includegraphics[width=1.0\textwidth]{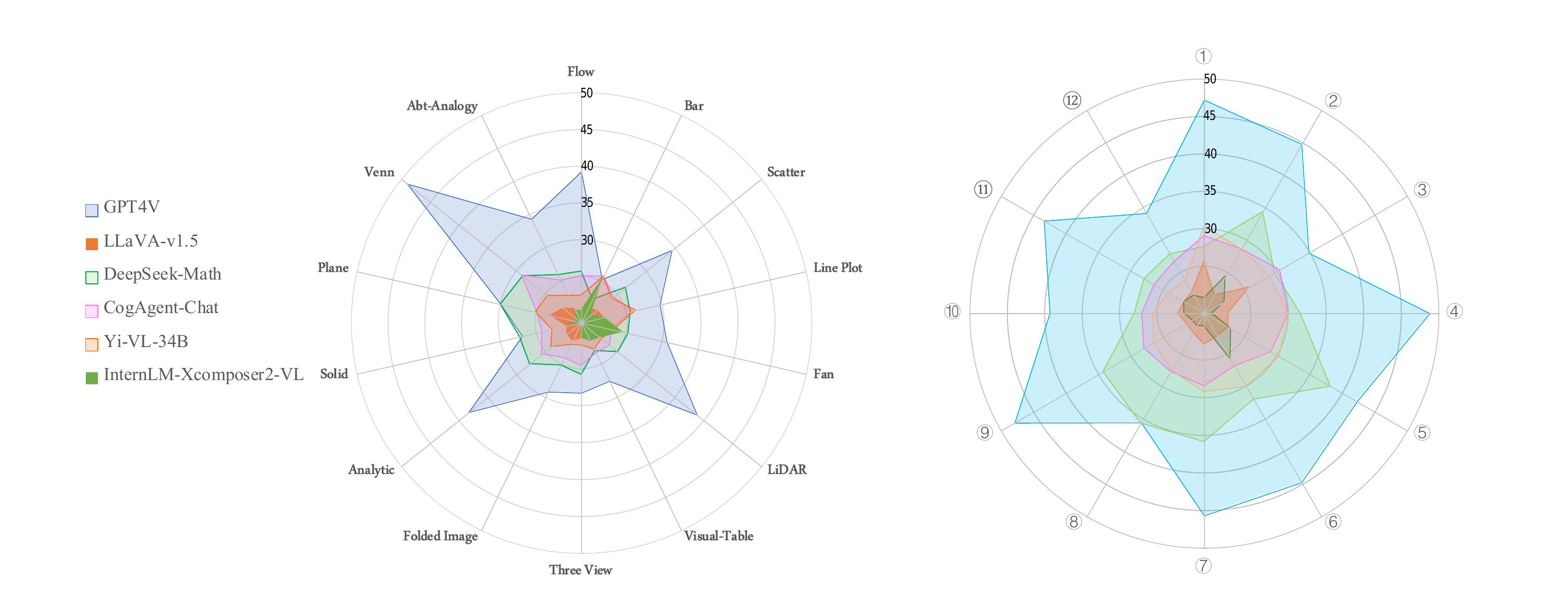}
  \caption{The results of mainstream multimodal large models and pure text large models on the CMMaTH dataset. \textbf{Left}: represents the performance evaluation of selected LMMs and LLMs across various Visual Subjects. \textbf{Right}: the performance assessment of these models on different educational grade-level questions.}
  \label{fig:teaser}
  \end{center}
  \vspace{-6mm}
\end{figure*}

Existing Math benchmarks for answer evaluation can be categorized into two types:\textit{Rule-based} \cite{GSM8K, MATH, OlympiadBench} and \textit{API-based} methods \cite{MathVista, GeoEval, MMLU}. \textit{API-based} methods are very costly and time-consuming, and they often result in unstable and inconsistent evaluation results. \textit{Rule-based} methods, on the other hand, struggle to handle highly diverse contents of benchmarks. Also, it is difficult to maintain handcrafted rules for dynamically updated benchmarks. Current multimodal math benchmark evaluations often resort to multiple-choice or true/false question formats, using rules or API-based LLM to extract options for assessing answers.

Based on above considerations, we propose a new multimodal mathematical benchmark CMMaTH.  Compared to previous benchmarks, our benchmark demonstrates greater diversity, increased depth of reasoning, and finer-grained knowledge annotation for multimodal models to grasp different levels and types of knowledge. We provided and open-sourced a lightweight answer comparator called GradeGPT, designed to compare the consistency between outputs from different LLM/LMMs and standard answers, thus avoiding expensive evaluation costs. Leveraging the CMMaTH dataset and GradeGPT tool, we evaluated mainstream open-source and commercial multimodal large models in Table \ref{tab:main_model_performance}, reporting comprehensive evaluation results along with extensive case analyses.
In summary, our paper makes the following contributions:
\begin{itemize}
\item We introduce the largest high-quality Chinese multimodal mathematics benchmark with the most detailed annotation granularity to date. We also provide an English version of this dataset. The CMMaTH dataset is a dynamically maintained and will be periodically updated.
\item Compared to previous multimodal mathematical benchmarks, our dataset exhibits great depth of reasoning and diversity. Our benchmark simulates more realistic educational Q\&A scenarios, encompassing a wider variety of question types and answer formats. Additionally, we annotate each question with detailed knowledge points and corresponding skills to evaluate the mastery level of current large models.
\item We build an evaluation assistant named GradeGPT on the CMMaTH dataset, which allows for comparing the proximity of model responses to standard answers and assessing the correctness of results and processes. GradeGPT features lightweight open-source characteristics, avoiding the instability and high costs associated with commercial models.
\item We conduct a systematic evaluation of existing mainstream multimodal large models, quantitatively and qualitatively comparing with existing models.
\end{itemize}

\section{Related Work}
\subsection{Assessment of mathematical abilities}
To evaluate the performance of large models in mathematical reasoning and examine hallucinations during the reasoning process, numerous benchmarks have been proposed for evaluating the mathematical reasoning capabilities of large models. GSM8K\cite{GSM8K} is the first and most widely used mathematical dataset used for large model math evaluation, consisting of 1k math word problem test samples and corresponding answers. The MATH\cite{MATH} dataset, in comparison to GSM8K, presents a greater challenge in terms of reasoning difficulty. This dataset demands a more profound understanding and intuition in various mathematical domains such as Algebra, Number Theory, and Geometry. MathVista\cite{MathVista} is the first dataset used to evaluate the multimodal mathematical capabilities of large models, but it has relatively simple reasoning depth. MATH-VISION\cite{MATH-VISION} has richer visual elements and deeper reasoning difficulty. MathVerse\cite{MathVerse} constructed several subsets of datasets to assess whether existing multimodal large models can truly understand mathematical abstract forms.

The CMMaTH Benchmark, in comparison to existing works on the evaluation of mathematical proficiency, places a greater emphasis on the analysis of mathematical abilities within the context of the Chinese language. The data distribution of the CMMaTH dataset more closely aligns with the actual distribution found in K12 educational settings, and it provides detailed annotations of mathematical knowledge points to facilitate the assessment of models’ mastery of knowledge and skills.
\subsection{Large Model Evaluation Tool}
Due to their strong generalization capabilities and extensive world knowledge, large language models have achieved outstanding results in tasks such as machine translation\cite{MachineTranslation}, question answering\cite{LLMQA}, dialogue\cite{LLMDialogues} and so on by generating text. Evaluating the comprehensive abilities of large models, such as clarity, adherence to instructions, comprehensiveness, formality, and mathematical reasoning ability, has received widespread attention\cite{CritiqueLLM}. Currently, many works opt to use powerful commercial model APIs, such as GPT-4, to assist in evaluating the comprehensive abilities of large models. For instance, MathVista\cite{MathVista} and GeoEval\cite{GeoEval} use GPT-4's API to extract correct answers for evaluation. These methods face several challenges: they are costly and time-consuming, and they struggle to keep up with rapid model iterations. Besides, these methods face challenges in terms of consistency and reproducibility\cite{LLMnotFair, CritiqueLLM}. 

Recent methods have proposed using metrics such as BERT score\cite{BERTScore} or MAUVE\cite{MAUVE} for evaluation. However, the numerical indicators produced by these methods are difficult to interpret when it comes to the erroneous responses generated by LLM. PandaLM and CritiqueLLM \cite{PandaLM, CritiqueLLM} are similar to our work. They proposed a fine-tuning method based on open-source LLMs, distilling the evaluation capabilities of GPT-3.5 into a series of smaller open-source models. However, they are focused on the automated evaluation of more general text generation tasks, while we are targeting the automated evaluation of responses from large models for multimodal mathematics problems.

Unlike PandaLM\cite{PandaLM} trying to evalution relative
conciseness, clarity and so on, our evaluation model, GradeGPT, is a dataset-oriented answer comparator that can provide specific reasons based on the standard answer and a model's response. We distilled the answer comparison capability of GPT-4 using the Cross-Lingual Judge-of-Chain method and enhanced GradeGPT's answer discrimination ability.

\section{CMMaTH Dataset}
\begin{figure}[t]
 \small
 \centering
 \begin{tabular}{lc}
         \toprule
         \textbf{Statistic} & \textbf{Number} \\
         \midrule
          Total questions & 23856 \\
          ~- multiple-choice questions & 18191 \\
          ~- Free-form questions & 5665 \\
          ~- Questions in the testmini set & 1000 \\
        \midrule
        Single-choice questions & 13706(75.3\%)\\
          ~- Proportion of answers A & 2694(14.8\%)\\
          ~- Proportion of answers B & 3903(21.4\%)\\
          ~- Proportion of answers C & 3961(21.7\%)\\
          ~- Proportion of answers D & 3148(17.5\%) \\
        Multiple-choice \& Multi-turn questions & 4485(24.7\%)\\
         \midrule
         knowledge point number & 2299 \\
         \midrule
         Levels & 5 \\
         Visual Subjects & 13 \\
         \midrule
         Maximum question length & 593 \\
         Minimum question length & 6 \\
         Average question length & 75.1 \\
         \midrule
         Grade Distribution
         Elementary(1-6) & 800 \\
         Junior(7-9) & 5082 \\
         Senior(10-12) & 17972 \\
         \bottomrule
     \end{tabular}
 \captionof{table}{Key statistics of CMMaTH. The unit of question length is words.}
 \label{tab:overview}
 \vspace{-5mm}
\end{figure}

\subsection{Overview of CMMaTH}
\begin{figure*}[t]
    \begin{center}
    \includegraphics[width=1.7\columnwidth]{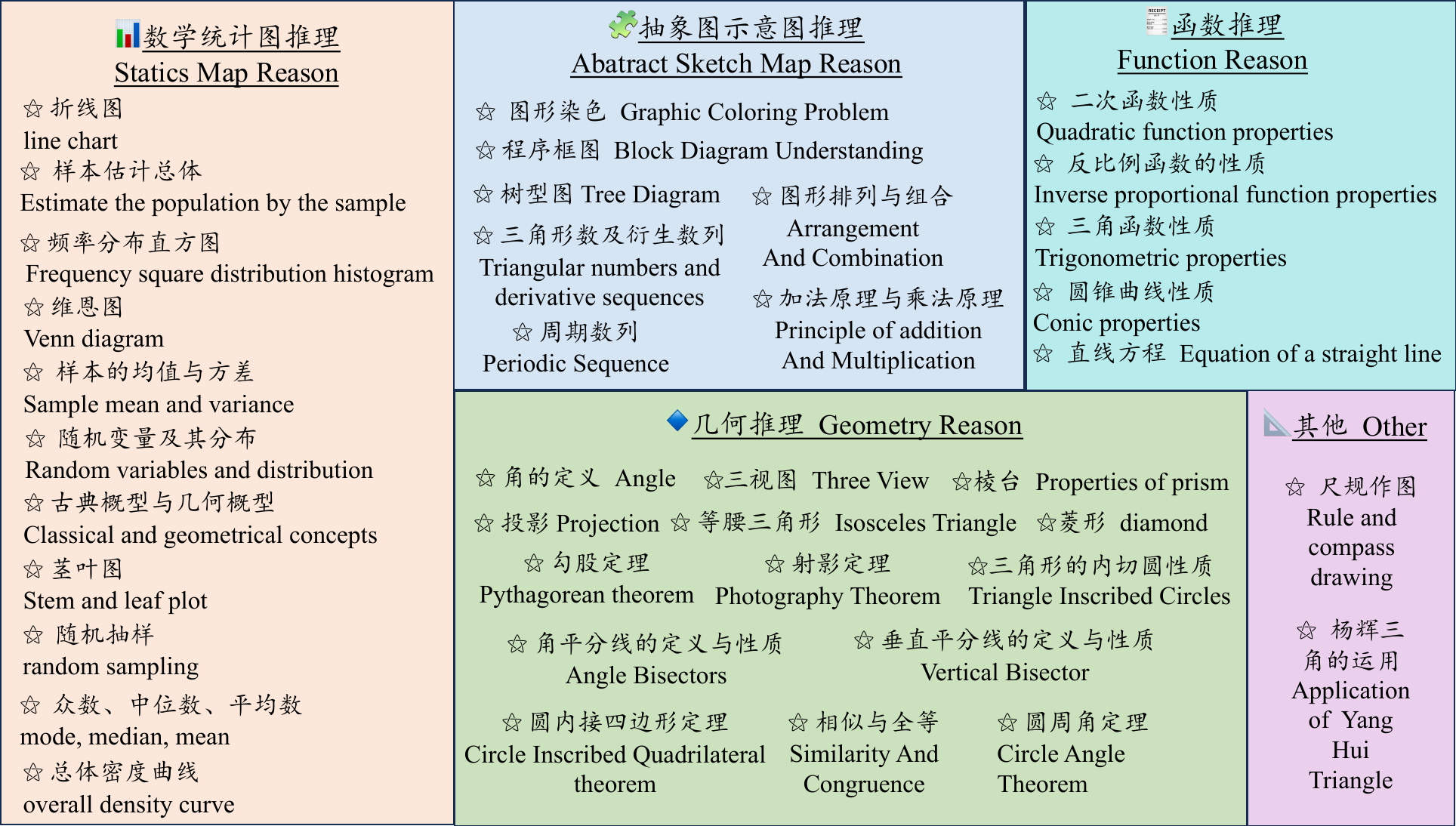} 
    \end{center}
    \caption{Some of the knowledge points involved in the CMMaTH dataset.}
    \label{fig:kn_overall}
    \vspace{-4mm}
\end{figure*}

We selected diverse multimodal mathematical problems from a vast pool of K12 educational questions, comprising 23856 items across 13 visual themes, 5 difficulty levels, and encompassing 150 types of knowledge points. More detailed statistical data can be found in Table \ref{tab:overview}.

For the convenience of evaluation, we provide a miniaturized test set of CMMaTH, called CMMaTH-testmin, containing 1500 samples. Testmin retains the diversity of the CMMaTH dataset and shows similar overall performance to the entire CMMaTH dataset. Evaluators can conduct quick tests and generate preliminary analyses based on CMMaTH-testmin.

\begin{CJK}{UTF8}{gbsn}
\begin{table*}[ht]
\centering
\small
\begin{tabular}{@{}cccccccc@{}}
\toprule
\textbf{Image Type} & \textbf{\#Num} & \textbf{Image Type} & \textbf{\#Num} & \textbf{Image Type} & \textbf{\#Num} & \textbf{Image Type} & \textbf{\#Num} \\
\midrule
\makecell{视觉表格\\ \textit{Visual-Table}} & 1513 & \makecell{折叠展开图\\ \textit{Folded Image Graph}} & 235 & \makecell{立体几何图\\ \textit{Solid Geometry}} & 2054 & \makecell{解析几何图\\ \textit{Analatic Geometry}} & 3060 \\
\midrule
\makecell{流程图\\ \textit{Flow Chart}} & 3120 & \makecell{条形图\\ \textit{Bar Chart}} & 4924 & \makecell{散点图\\ \textit{Scatter Chart}} & 517 & \makecell{平面几何图\\ \textit{Plane Chart}} & 3834 \\
\midrule
\makecell{折线图\\ \textit{Line Chart}} & 846 & \makecell{饼状图\\ \textit{Fan Chart}} & 175 & \makecell{雷达图\\ \textit{LiDAR Chart}} & 73 & \makecell{抽象类比图\\ \textit{Abatract Analog Graph}} & 440 \\
\midrule
\makecell{三视图\\ \textit{Three View Graph}} & 22 & \makecell{枝页图\\ \textit{Stem-and-Leaf display}} & 23 & \multicolumn{3}{c}{\makecell{其他\\\textit{Other Image type}}} & 240  \\
\bottomrule
\end{tabular}
\caption{Primary element types involved in the CMMaTH dataset.}
\label{tab:images_type}
\vspace{-5mm}
\end{table*}
\end{CJK}
\subsection{Collectioin Guidelines}
We collected a large number of multimodal mathematics questions from a vast K12 educational question bank, including elements such as statistical charts, plane geometry, three-view diagrams, flowcharts, set notation diagrams, etc. The quality and distribution of the data were guided by the following criteria during collection.

\begin{itemize}
\item Diverse Mathematical Visual Elements. We have collected solutions to multimodal mathematical problems that rely on understanding image content, especially those containing a large amount of Chinese visual content such as text and symbols. Table \ref{tab:images_type} shows some visual elements subject of CMMaTH.
\item High relevance to the K12 math knowledge and skill. The annotator, who is well-versed in knowledge, needs to ensure that the multimodal question assesses a specific K-12 mathematics knowledge point during the question collection process. It primarily includes mathematics questions related to K12 education, facilitating the assessment of the application potential of large-scale multimodal capabilities in the field of mathematics education. 
\item High-quality images and answers. During the collection phase, we instruct collectors to disregard multimodal math questions with erroneous symbols or low-quality images (blurry images). Collectors are required to ensure that the collected questions are generally solvable.

\end{itemize}
\subsection{Data Collections}
\textbf{Collection from Diverse Multimodal Math Sources} CMMaTH's data is based on a million-level private database. The private database we used comes from questions collected from the Internet and undergoes rigorous data checking. The project’s data has undergone multiple rounds of collection. We first sampled 45,000 multimodal math questions: 14,000 each from elementary, high, and junior high schools. Then, we added 34,000 more questions featuring algorithm block diagrams, statistics, and geometry diagrams to enhance visual diversity. \\
\textbf{Data Filtering} We filtered out all questions without images in the question stems, including questions with multi-graph reasoning, questions in non-Chinese languages, and questions not relying on visual content to solve. To ensure the quality of the images and text questions, we removed all images whose width and height were less than 100, then used the GPT4 API to score the data quality and filter out questions suspected of being unsolvable and questions with garbled text in the question text. \\
\textbf{Data Labeling} For K-12 mathematics knowledge points, we have scraped the mathematics section from Jiaoyan Cloud\footnote{\url{https://www.jiaoyanyun.com/}} and organized all the knowledge points into a knowledge tree including a total of 5,531 knowledge points. We retained 2,299 knowledge points more relevant to multimodal mathematics in K-12. Subsequently, all questions were classified according to knowledge points by GPT-4 and a fine-tuned LLM, followed by manual multi-level verification. Questions that did not match any K-12 multimodal mathematics knowledge points were filtered out. \\
\vspace{-5mm}

\subsection{Comparison with Existing Benchmarks}
\begin{table*}
\centering
\resizebox{2.0\columnwidth}{!}
{
\begin{tabular}{@{}lcccccccc@{}}
  \toprule
  Dataset & Size & Image\&Supplementary Input & Format & Source & Answer & Knowledge Annotation & Lanugage Domain & Knowledge Domain\\
  \midrule
  VQAv2\cite{VQAv2} & $>1$M & V & I+T & Annotated & Open/MC/TF & \textcolor{red}{\text{\XSolidBrush}} & En & General \\
  SEED\cite{SEED-Bench}  & 19K & V & I+T & Annotated & MC & \textcolor{red}{\text{\XSolidBrush}} & En & General \\
  MMBench\cite{MMBench} & 3K & V & I+T & Repurposed & MC & \textcolor{red}{\text{\XSolidBrush}} & En & General \\
  MM-Vet\cite{MMVet} & 0.2K & V & I+T & Annotated & Open & \textcolor{red}{\text{\XSolidBrush}} & En & General \\
  ScienceQA\cite{ScienceQA} & 6K & V & I+T & Textbooks & MC & \textcolor{red}{\text{\XSolidBrush}} & En & Science \\
  MathVista\cite{MathVista}  & 1K/6K & V(5 Types)+OC & I+T & Synthesized & Open/MC/TF & \textcolor{red}{\text{\XSolidBrush}} & En/ZH & Math \\
  MMMU\cite{MMMU}  & 11.5K & V(30 Types)+OC & Interleaved & Textbooks & Open/MC & \textcolor{red}{\text{\XSolidBrush}} & -- & General \\
  CMMMU\cite{CMMMU}  & $<1$K(Math Part) & V(5 Types)+OC & Interleaved & Internet & Open/MC & \textcolor{red}{\text{\XSolidBrush}} & ZH & General\\
  OlympiadBench\cite{OlympiadBench} & 6.5K(Math Part) & V(5 Types) & Interleaved & Internet & Open & \textcolor{red}{\text{\XSolidBrush}} & ZH/EH & Math/Physics \\
  MathVerse\cite{MathVerse} & 2.6K/15K & V(3 Types) & I+T & Synthesized & MC & \textcolor{red}{\text{\XSolidBrush}} & ZH/EH & Math \\
  MATH-Vision\cite{MATH-VISION} & 3K & V(16 Types)+IC & I+T & Synthesized & Open/MC & \textcolor{red}{\text{\XSolidBrush}} & EH & Math \\
  \midrule
  CMMaTH & 23K & V(13 Types)+OC+IC & I+T &  Internet/Annotated & Open/MC/TF & \textcolor{green}{\checkmark} & ZH & K12 Math \\
  \bottomrule
\end{tabular}
}
\caption{Comparison with other multimodal benchmarks. V: visual input, VD: video input, OC: optical characters, IC: Image Caption, I+T: images and text, Open: open questions, MC: multiple choice questions, FIB: fill in the blank questions, TF: true or false questions.}\label{tab:comparison with multimodal benchmarks}
\vspace{-0.2cm}
\end{table*}
The CMMaTH dataset is primarily used to evaluate multimodal reasoning capabilities in K-12 educational scenarios. We compared the current mainstream multimodal mathematical datasets and large model benchmarks in Table \ref{tab:comparison with multimodal benchmarks}. Compared to existing multimodal benchmarks and multimodal reasoning benchmarks, the CMMaTH dataset has the following characteristics: \\
\textit{\textbf{Extreme Diversity}} Currently, there is a severe lack of high-quality Chinese multimodal mathematics datasets. MATH-VISION lacks a Chinese component, the MATH-VISTA dataset contains only a small number of Chinese samples, and CMMMU contains only 540 math problems, which are not fine-grained and comprehensive enough. We have included about 23k fine-grained multimodal mathematics assessment samples, covering 13 K12 mathematics visual categories, making it the largest known multimodal Chinese dataset to date. \\
\textit{\textbf{Real and High Quality \& Multilingual}} MathVista features a substantial number of problems that are associated with natural and synthetic images. However, these images do not accurately represent the genuine data distribution encountered in K12 mathematics educational settings. OlympiadBench is an Olympiad-level bilingual multimodal benchmark. However, this benchmark is overly challenging and deviates from the application of LMM in real  K12 multimodal math scenarios. Additionally, the variety of multimodal visual elements is relatively limited. Instead, we collect multimodal data specifically tailored to the K12 education context. Additionally, MathVista incorporates a significant amount of data from GeoQA and synthetic images, which have relatively poor image quality. Our multimodal visual image elements have all undergone stringent image quality assessments. Unlike CMMMU, CEval, and CMath, our dataset is a bilingual dataset that considers a large number of Chinese scenes. In addition to the text of the questions being in Chinese, the visual elements related to the questions also contain Chinese text/symbols. \\
\textit{\textbf{High-quality Fine-grained Annotation and Evaluation Tool}} Every question in our dataset is meticulously annotated with standardized answers, solutions expressed in natural language, associated multimodal knowledge points, visual element categories, and K-12 grade levels. This fine-grained annotation enables a more nuanced evaluation of multimodal mathematical proficiency within the K-12 educational context. While MathVista and GeoEval rely on GPT-4 for answer extraction and validation, we introduce an open-source model named GradeGPT. GradeGPT stands out by providing a stable, cost-free, and swift accuracy evaluation specifically tailored for the CMMaTH dataset.

\begin{figure*}[t]
    \begin{center}
    \includegraphics[width=1.8\columnwidth]{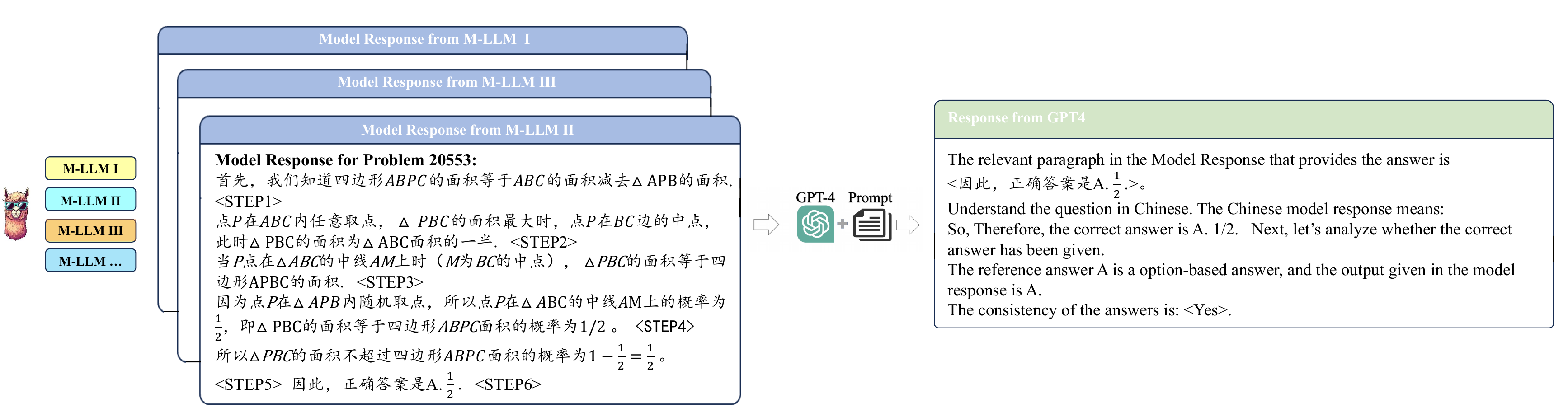} 
    \end{center}
    \caption{Instruction Construction Pipeline of GradeGPT}
    \label{fig:gradegpt_instruction}
    \vspace{-5mm}
\end{figure*}

\section{GradeGPT}
The CMMaTH dataset encompasses a large variety of problem-solving objectives, such as mathematical expressions, multiple-choice options, numerical outcomes, coordinate points, conclusion figures, and correctness assessments. Traditionally, in reasoning or evaluation contexts, problems have been formulated as multiple-choice or true/false questions to facilitate comparison and to simplify the extraction of results. Also, it is difficult to maintain dynamically updated benchmark. Employing API models for evaluation is prohibitively expensive, and the resulting evaluations are not consistently stable, which also hampers the iterative development of models on benchmarks, such as hyperparameter selection. 

To provide a stable, free, fast, and easy-to-update model response evaluation tool, we introduce GradeGPT, an answer comparison model tailored for the CMMaTH dataset. GradeGPT is designed to receive a question, its standard answers, and a model-generated response. It extracts key steps including results from Chinese output. Determine whether the result is consistent with the standard answer. Our GradeGPT is a streamlined, open-source model. When integrated with frameworks such as vLLM using the 14B model, it can swiftly compare a myriad of model-generated answers, accomplishing a remarkable judgment accuracy of 96.1\% for assessing responses comparable with GPT4 API. \\
\textbf{Prompt Format} \\
In the prompt input of GradeGPT, there are "questions," "reference answers," and "model output answers." The model is required to provide an answer in the form of "<Yes>" or "<No>" indicating whether the model output answer is equivalent to the standard reference answer. We have designed an instruction format named Cross-Lingual-Judge-of-Chain for the purpose of determining answer consistency. Cross-Lingual-Judge-of-Chain first analyzes the model response and finds the key sentences that give the answer in the model response, understand key chinese sentences in English. Then analyze the standard answer, determine the type of the standard answer, and then determine whether the standard answer is included in the model response. More details can be found in Appendix \ref{GradeGPT_detail} \\
\textbf{Instruction Construction} \\
We first generate inference results on CMMaTH using multiple Multimodal LLMs and provide GPT-4 with a detailed few-shot prompt to synthesize answer judgments in the form of a Cross-Lingual Judge-of-Chain response. By employing GPT4’s In-Context Learning, as showned in Figure \ref{fig:gradegpt_instruction}, we have established a procedure for synthesizing instruction data and have produced approximately 56k cross-lingual result judge instruction pairs. Through fine-tuning the model with these instructions, we are able to obtain an expert model, GradeGPT, which possesses the capability to compare answers.
 \\
\vspace{-4mm}
\section{Experiments}
\vspace{-2mm}
We conducted a series of experiment to evaluate various models on the CMMaTH dataset. We evaluated various LLM/LMM models, including open-source and closed-source models. More model details can be found in Table \ref{tab:model-hyperparameters}. We employed a method similar to GeoEval and MathVista, generating captions through an GPT4V, and assessed them using MetaMath, and DeepSeekMath equipped with caption information. Our empirical research reveals that even the most advanced models struggle to achieve satisfactory accuracy levels. Furthermore, we conducted an exhaustive error analysis on a sufficiently strong commercial multimodal model, GPT-4V, examining its error distribution and presenting illustrative qualitative examples. Our investigation also revealed that the inclusion of multilingual thought chains does not mitigate the substantial difficulties presented by Chinese multimodal mathematical reasoning scenarios. We postulate that the richness of non-English contextual information contained within the images necessitates models equipped with enhanced multilingual OCR and sophisticated multimodal diagram reasoning capabilities.

\subsection{Main Experiments on LLM/LMMs}

\begin{figure}[t]
    \begin{center}
    \includegraphics[width=1.0\columnwidth]{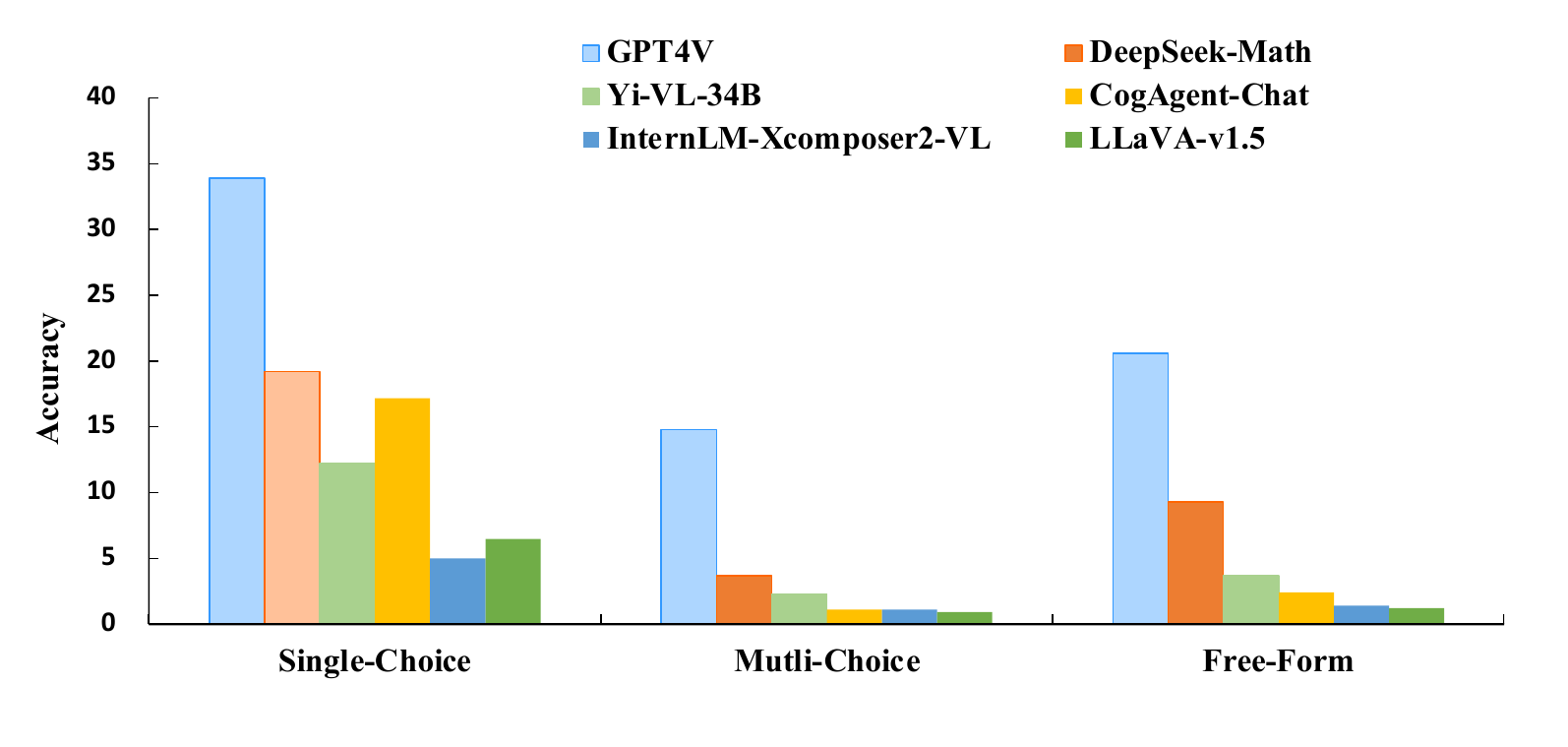} 
    \end{center}
    \caption{Accuracy of LMMs across different types of problems in CMMaTH Benchmark.}
    \label{fig:question_type_acc_bar}
\end{figure}

\begin{figure}[t]
    \begin{center}
    \includegraphics[width=1.0\columnwidth]{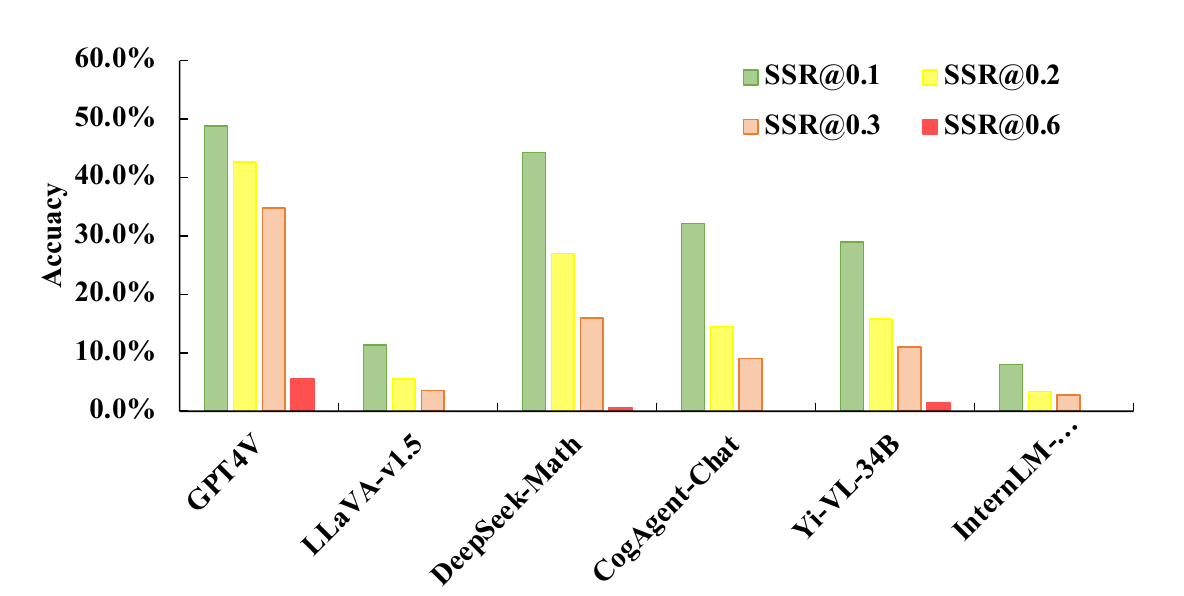} 
    \end{center}
    \caption{The metrics of different LMMs/LLMs models about SSR.}
    \vspace{-2mm}
    \label{fig:ssr_rate_bar}
\end{figure}

\begin{table*}[htbp]
\centering
\resizebox{\textwidth}{!}{%
\begin{tabular}{l|c|cccccccccccccc}
\toprule
Model & Overall & Flow  & Bar & Scatter & Line Plot & Fan & LiDAR  & Visual-Table & Three View & Folded Image & Analytic  & Solid & Plane & Venn & Abt-Analogy  \\
\midrule

\multicolumn{16}{c}{LLMs (Text Only)}\\
\midrule
LLama2-70B & 4.5 & 4.7 & 2.5 & 4.4 & 7.9 & 7.4 & 8.1 & 3.4 & 5.4 & 5.1 & 5.3 & 4.1 & 5.3 & 5.9 & 4.5 \\
MetaMath-70B & 5.7 & 4.6 & 3.3 & 6.6 & 8.7 & 5.7 & 0.2 & 4.2 & 4.1 & 8.5 & 7.2 & 4.8 & 8.5 & 9.8 & 5.4 \\
DeepSeek-Math & 14.0 & 13.4 & 6.7 & 14.7 & 13.1 & 12.5 & 12.2 & 8.1 & 13.5 & 12.3 & 17.2 & 16.5 & 21.6 & 19.5 & 13.8 \\
Baichuan-13B & 8.4 & 6.7 & 4.8 & 12.2 & 12.4 & 13.1 & 16.2 & 5.4 & 4.1 & 8.5 & 11.1 & 6.7 & 13.7 & 12.8 & 9.3 \\
Qwen-14B & 13.7 & 15.5 & 7.3 & 14.3 & 16.9 & 13.6 & 10.8 & 11.4 & 12.8 & 14.8 & 15.9 & 12.7 & 17.8 & 20.4 & 19.3 \\
\midrule

\multicolumn{16}{c}{Math LLMs (Text + OCR Caption)}\\
\midrule
LLama2-70B & 5.6 & 4.9 & 2.3 & 4.8 & 7.9 & 7.1 & 8.0 & 4.4 & 6.4 & 9.1 & 3.3 & 4.8 & 6.3 & 6.9 & 5.5 \\
MetaMath-70B & 5.1 & 4.3 & 3.2 & 6.9 & 8.1 & 5.3 & 0.0 & 4.4 & 4.2 & 8.8 & 7.1 & 4.4 & 8.3 & 9.1 & 5.2 \\
DeepSeek-Math & 15.3 & 13.2 & 6.9 & 14.1 & 12.6 & 12.3 & 12.1 & 8.9 & 14.4 & 14.1 & 17.9 & 19.3 & 22.7 & 21.5 & 13.9 \\
Baichuan-13B & 8.1 & 6.9 & 4.3 & 12.4 & 11.5 & 12.3 & 14.9 & 3.4 & 4.4 & 9.3 & 11.6 & 6.8 & 13.2 & 12.9 & 9.9 \\
Qwen-14B & 13.3 & 14.1 & 7.4 & 13.3 & 16.2 & 13.2 & 11.8 & 10.6 & 11.8 & 19.8 & 5.9 & 11.7 & 13.8 & 21.4 & 16.3 \\
\midrule

\multicolumn{16}{c}{Open-source LMMs (Text + Image)}\\
\midrule
LLaVA-v1.5-7B & 5.5 & 1.5 & 4.2 & 5.4 & 6.2 & 5.4 & 3.6 & 4.0 & 4.2 & 5.3 & 4.8 & 3.9 & 8.4 & 6.1 & 4.2   \\
InternLM-XComposer2-VL & 3.4 & 3.3 & 5.3 & 3.2 & 6.2 & 11.3 & 6.2 & 5.4 & 4.0 & 0.5 & 0.4 & 3.6 & 1.5 & 1.8 & 3.6  \\
Yi-VL-34B & 8.3 & 7.1 & 4.6 & 10.2 & 14.6 & 8.5 & 6.8 & 7.7 & 5.9 & 6.4 & 10.1 & 7.8 & 12.2 & 11.3 & 7.9  \\
CogAgent-Chat & 10.6 & 12.2 & 5.2 & 10.8 & 13.7 & 8.0 & 9.5 & 8.8 & 11.2 & 10.2 & 13.2 & 10.5 & 11.8 & 19.9 & 12.2 \\

\midrule
\multicolumn{16}{c}{Closed-source LMMs (Text + Image)}\\
\midrule
GPT4V & 27.0 & 39.3 & 12.5 & 30.2 & 21.0 & 22.9 & 38.6 & 16.9 & 18.3 & 20.0 & 37.5 & 15.8 & 21.5 & 
58.0 & 29.9\\

GPT4o & 35.2 & 59.4 & 18.8 & 54.5 & 31.7 & 58.4 & 32.4 & 31.7 & 28.7 & 23.8 & 40.6 & 31.6 & 33.6 & 
57.4 & 29.7\\
\midrule
\multicolumn{16}{c}{Human Performance}\\
\midrule
Human (testmini) & 80.1 & 73.7 & 78.9 & 96.2 & 95.1 & 57.4 & 91.7 & 83.5 & 69.2 & 63.2 & 67.5 & 51.6 & 72.1 & 89.1 & 83.1\\

\bottomrule
\end{tabular}}
\caption{Comparison of model performances across various mathematical subjects. Subjects: Flow: Flow Chart, Bar: Bar Chart, Scatter: Scatter Chart, Line Plot: Line Curve and Plot, Fan: Fan Chart, LiDAR: LiDAR Chart, Visual-Table: Visual-Table Chart, Three View: Three View Graph, Folded Image: Folded Image Graph, Analytic: Analytic Geometry Problem, Solid: Solid Geometry Problem, Plane: Plane Geometry Problem, SolG: Venn: Set Venn Graph, Abt-Analogy: Abtract Analogy Graph.}
\label{tab:main_model_performance}
\vspace{-3mm}
\end{table*} 
We evaluated the results of mainstream multimodal large models and mathematical expert models in Table \ref{model_detail}. We analyzed the trend of existing large models in descending with problems and conditions, as well as the effectiveness of techniques such as Cross-Lingual Prompting in solving Chinese multimodal mathematical problems. The experimental in Table \ref{tab:main_model_performance} results indicates that our data exhibits extremely strong diversity and relatively challenging reasoning depth. Figure \ref{fig:teaser} and Table \ref{tab:main_model_performance} shows models such as GPT4V struggle to comprehend our multimodal content and reasoning questions effectively, resulting in significant performance gaps between open-source and proprietary models. In certain rare visual domains, multimodal large models achieve very low reasoning outcomes. \\
\textbf{Accuracy on various question types.} We evaluated the accuracy of GPT4V on various target-solving tasks in Figure \ref{fig:question_type_acc_bar}. The results indicate that when solving free-form problems, especially those with more diverse targets such as expressions, coordinates, and conclusion judgments, the multimodal large language model shows poorer performance. \\
\textbf{Is OCR information sufficient for CMMaTH?} We also referred to works like MathVista, attempting to use LLMs combined with OCR information from diagrams to assist in mathematical reasoning in Table \ref{tab:main_model_performance}. We found that, in our benchmark, a small amount of OCR information (such as mathematical symbols in diagrams, axis values, and image titles) made it very difficult to complete our multimodal mathematical reasoning tasks. The results indicate that solving problems in CMMaTH requires stronger multimodal mathematical chart capabilities, beyond just OCR. \\
\textbf{K12 Multimodal Knowledge Richness of current LMMs.} We systematically evaluated the proficiency of existing multimodal large models in the K12 domain regarding multimodal reasoning skills in Figure \ref{fig:ssr_rate_bar}. The results revealed a significant knowledge gap in existing multimodal K12 educational resources. Compared to other existing LMMs, GPT4V possesses a richer knowledge base, thereby substantially reducing the illusion of reasoning in multimodal mathematical inference.
\vspace{-3mm}
\subsection{Experiments of Cross-language Reason Technology} 
We also attempted several multilingual Chain-of-Thought approaches such as En-CoT, CLP(Cross-Lingual Prompting) used by \citet{CLP2023} to observe whether multimodal mathematical problems could be enhanced through context learning techniques without training. The results indicate that multilingual CoT methods face challenges in solving, possibly due to the abundance of Chinese contextual text in the image content, which may necessitate the model to demonstrate excellent cross-lingual OCR capabilities. We have included more details on the implementation of Cross-Lingual Prompting and En-CoT on the CMMaTH dataset in the Table \ref{tab:cross_lin_tech}. 

\begin{table}
    \small
    \label{Aba:Cross-Ling Experiment}
    \centering
    \begin{tabular}{lcc}
        \toprule
        LMM   & Overall-Acc  \\
        \midrule
        LLaVA-v15 & 4.2 \\
        InternLM-XComposer2-VL & 3.4 \\
        \midrule
        LLaVA-v15 + En-CoT &  9.4  \\
        InternLM-XComposer2-VL + En-CoT & 16.9 \\
        \midrule
        LLaVA-v15 + CLP &  12.7  \\
        InternLM-XComposer2-VL + CLP & 17.1 \\
        \bottomrule
    \end{tabular}
    \caption{The performance of train-free CoT reasoning techniques on the CMMaTH dataset.}
    \vspace{-2mm}
    \label{tab:cross_lin_tech}
\end{table}

\subsection{Error Analysis}
\begin{figure}[t]
    \begin{center}
    \includegraphics[width=1.0\columnwidth]{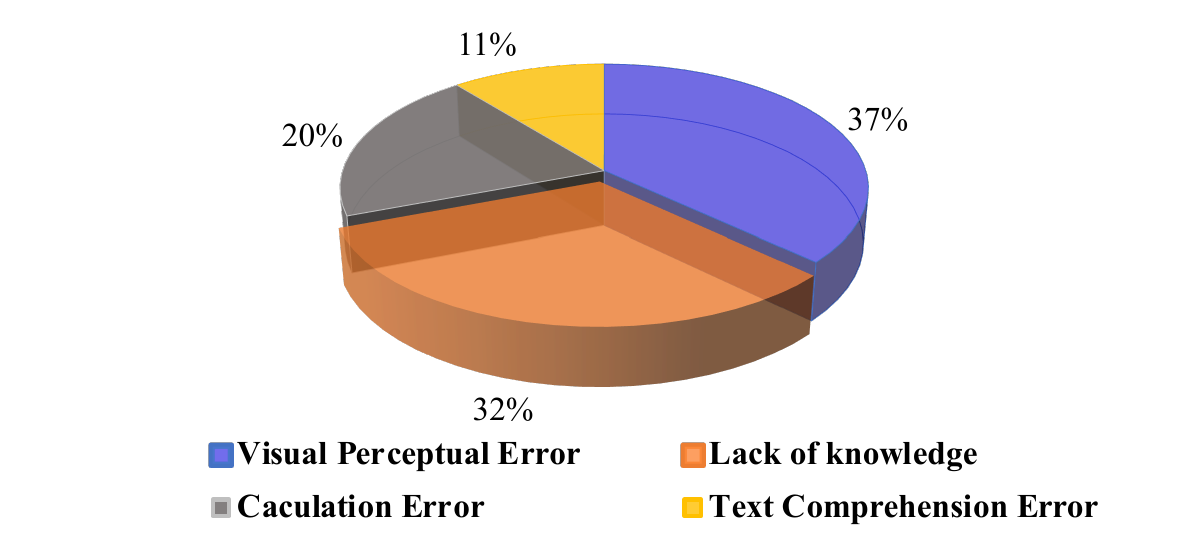} 
    \end{center}
    \caption{Distribution of Error Types in GPT4V.}
    \label{fig:error_type_bar_chart_statics}
    \vspace{-5mm}
\end{figure}
We conducted a detailed analysis and evaluation of GPT4V on CMMaTH-testmin, categorizing errors into four types: perceptual errors, reasoning errors, calculation errors, and Reject Errors. The error type distribution of GPT4V on CMMaTH is shown in the Figure \ref{fig:error_type_bar_chart_statics}. \\
\textbf{Perception Errors} \\
Perception Error refers to the model's erroneous interpretation and utilization of diagram content during reasoning. For example, incorrect OCR, misidentification of numerical relationships, geometric relationships, logical relationships, etc. \\
\textbf{Reasoning Errors}  \\
Reasoning Error are quite common during the solving process. For instance, the model may misinterpret symbols or use incorrect logic or knowledge for inference. The frequency of Reasoning Errors reflects the model's logical and mathematical reasoning capabilities. \\
\textbf{Calculation Errors} \\ 
Calculation Error refers to the model performing incorrect mathematical operations, such as writing equations or solving equations incorrectly. \\
\textbf{Reject Errors} \\
Reject Error refers to the model's inability to solve a problem that is actually solvable. The frequency of such errors reflects the model's ability to follow instructions. \\
\vspace{-3mm}
\section{Conclusions}

We introduce CMMaTH, a detailed Chinese math reasoning benchmark with diverse question types, vivid visuals, and complex reasoning. The benchmark includes detailed knowledge points, standard thought processes, and grade levels to measure the mastery of knowledge points in the K-12 multimodal math skill. To evaluate large multimodal models quickly and affordably, we built GradeGPT, an open-source tool for assessing results. Extensive experimental results on CMMaTH manifest the limitations of current models in multilingual, multimodal math reasoning.

\section*{Limitation \& Potential Impact}
Our dataset CMMaTH, as a multimodal mathematics dataset aimed at the K-12 education sector, can facilitate model evaluation and iteration of multimodal large models in this field, and may promote the research and development of educational artificial intelligence. CMMaTH primarily consists of single-image problems, without considering multi-image contextual reasoning or scenarios requiring auxiliary line drawing and similar tasks. GradeGPT is a result-oriented, relatively coarse reasoning response evaluator. How to construct a process evaluation model for fine-grained assessment of the reasoning ability of large models can continue to be explored in the future.

\bibliography{acl_latex}

\appendix
\section{More Related Work About Multimodal Large Model Evaluation} 
The multimodal large models face serious hallucination issues in perceiving objects and executing inference. To systematically evaluate the various capabilities of multimodal large models, diverse multimodal benchmarks are utilized for assessing the abilities of large models and aiding iterative development. POPE\cite{POPE} is used to evaluate the accuracy of large models in identifying perceptual objects. MMMU and CMMMU\cite{MMMU, CMMMU} are comprehensive subject datasets designed to assess the proficiency of large models in mastering massive multimodal multi-disciplinary knowledge. SEED-Bench designed 19,000 diverse multimodal questions spanning video and image modalities to evaluate the spatiotemporal capabilities of multimodal large models \cite{SEED-Bench}. MMVet\cite{MMVet} attempts to design datasets to evaluate the integrated capabilities of different multimodal large model systems in combining various Vision-Language skills.

\section{Model Generation Details}
\label{model_detail}
\subsection{Model Weight Version}
We evaluated models on CMMaTH, including open-source models such as  LLaVA-v1.5, Deepseek-Math, InternLM-XComposer2-VL, Yi-VL-34B, CogAgent-Chat, MetaMath-70B, LLama-70B, Baichuan-13B and Qwen-14B as well as state-of-the-art commercial models GPT4V. We have listed the parameter versions and the Hugging Face repository names of the open-source models used in Table \ref{tab:model_huggingface_version}.
\subsection{Model Sampling Parameter}
We have listed the corresponding hyperparameters used by the models in Table \ref{tab:model-hyperparameters}. For API models, we have indicated the corresponding release versions. Models using vLLM for inference are annotated.

\subsection{Data quality control} To ensure the high quality of the final data, we conducted sampling and manual verification. We performed three random samples, each consisting of 500 multimodal samples, to check the data quality and ensure the consistency of the knowledge points and data.

\section{Prompt Details}
\begin{figure*}[t]
    \begin{center}
    \includegraphics[width=2.0\columnwidth]{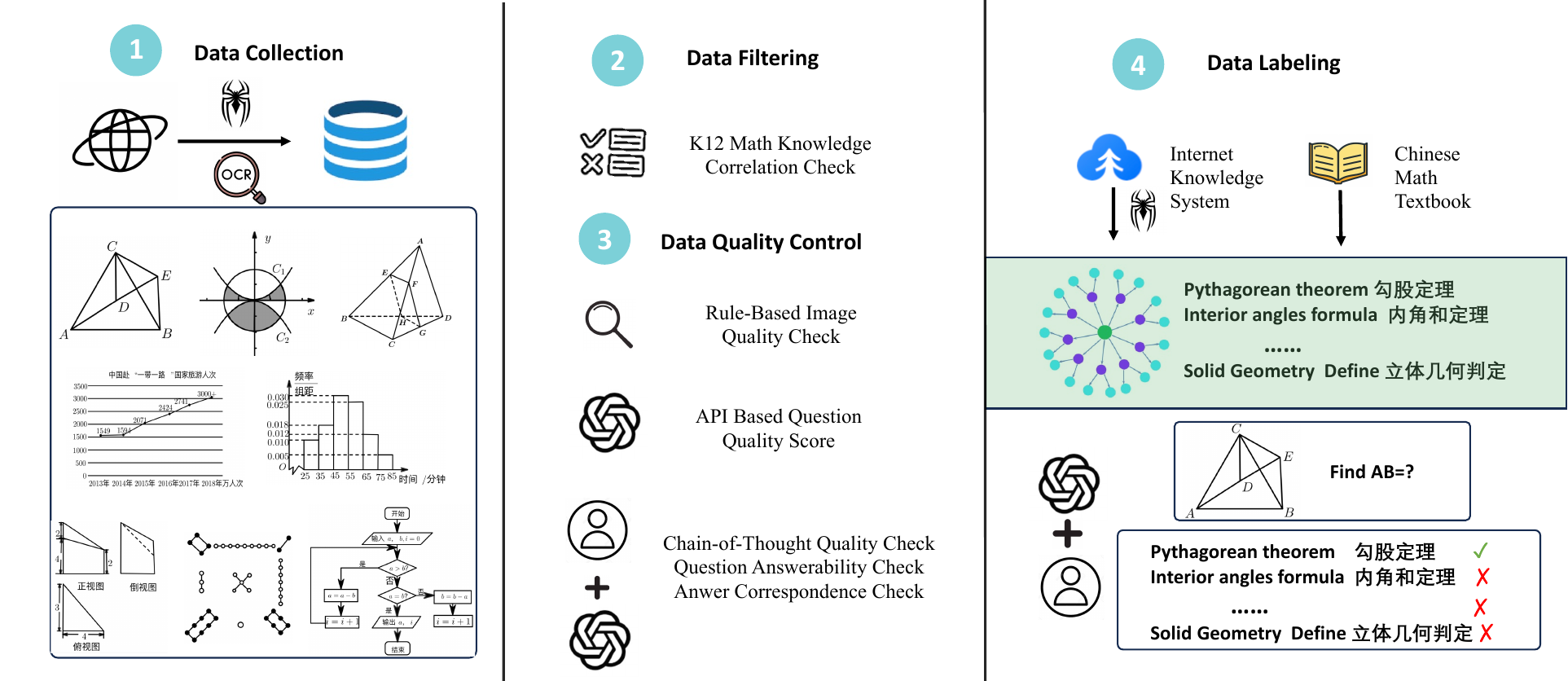} 
    \end{center}
    \caption{Overall Data Collection Pipeline of CMMaTH.}
    \label{fig:data_collection}
\end{figure*}
\subsection{Prompt For Step Response Generation}
\begin{table*}[tbhp!]
\begin{small}
\centering
\renewcommand\tabcolsep{2.5pt} %
\begin{tabular}{cp{13cm}}
\toprule
\textbf{Element} & \textbf{Prompt} \\
\midrule
system-prompt & 
\begin{minipage}[s][0.6cm]{1.5\columnwidth}
你是一个数学题解题助手，你的输入是一道数学题，以及这道题目的图像，你的任务是输出这道题目的求解思路以及答案. 输出格式是一步一步的方式.
\end{minipage}
\\
 
\midrule
Example1 & 
\begin{minipage}[s][1.0cm]{1.5\columnwidth}
\textbf{Question:} 小明有2000元, 一个苹果2元, 小明买了10个苹果。小明还有多少钱。
\textbf{Model response:} 首先计算小明买苹果的开销为2乘10等于20元. <STEP1> 用2000元减去所有的开销, 2000-20=1980. <STEP2>  所以最终的答案为1980元. <STEP3>
\end{minipage}
\\

\midrule
Example2 & 
\begin{minipage}[s][6.8cm]{1.5\columnwidth}
\textbf{Question:} 如图，在△ABC中，∠C=90°，∠B=30°，以A为圆心，任意长为半径画弧分别交AB、AC于点M和N，再分别以M、N为圆心，大于MN/2的长为半径画弧，两弧交于点P，连结AP并延长交BC于点D，则下列说法中正确的个数是（ ）①AD是∠BAC的平分线；②∠ADC=60°；③点D在AB的中垂线上；④BD=2CD．A.4 B.3 C.2 D.1 \\
\textbf{Model response:} 首先确定△ABC的几何性质。 <STEP1>∠C=90°，∠B=30°，∠A=120°。 <STEP2>△ABC是一个等腰三角形。 <STEP3>根据等腰三角形的性质，AD是∠BAC的平分线。 <STEP4>①AD是∠BAC的平分线，正确。 <STEP5>②∠ADC=60°，错误。 <STEP6>③点D在AB的中垂线上，错误。 <STEP7>④BD=2CD，正确。 <STEP8>因此，正确的个数为4。 <STEP9>答案为A.4。 <STEP10>请你根据这个例子，解决下面的数学题。问题：在△ABC中，∠C=90°，∠B=30°，以A为圆心，任意长为半径画弧分别交AB、AC于点M和N，再分别以M、N为圆心，大于MN/2的长为半径画弧，两弧交于点P，连结AP并延长交BC于点D，则下列说法中正确的个数是（ ）①AD是∠BAC的平分线；②∠ADC=60°；③点D在AB的中垂线上；④BD=2CD．A.4 B.3 C.2 D.1 求解步骤：首先确定△ABC的几何性质。∠C=90°，∠B=30°，∠A=120°。△ABC是一个等腰三角形。根据等腰三角形的性质，AD是∠BAC的平分线。①AD是∠BAC的平分线，正确。②∠ADC=60°，错误。③点D在AB的中垂线上，错误。④BD=2CD，正确。因此，正确的个数为4。答案为A.4。
\end{minipage}
\\
\bottomrule
\end{tabular} 
\end{small}
\captionsetup{justification=centering}
\caption{Prompt for all model to generate step-by-step answer.}
\label{tab:promt_for_step_generation}
\end{table*}
When evaluating hallucinations during the assessment process, we use a few-shot prompt format to elicit step-by-step outputs from the model as showed in Table \ref{tab:promt_for_step_generation}.

\subsection{Prompt For GradeGPT}
\begin{table*}[tbhp!]
\centering
\small
\renewcommand\tabcolsep{2.5pt} 
\begin{tabular}{cp{13cm}}
\toprule
\textbf{Element} & \textbf{Prompt} \\
\midrule
system-prompt & 
\begin{minipage}[s][2.6cm]{1.5\columnwidth}
You are an answer evaluation assistant. Based on the following questions and reference answers, determine if the model's output is the same answer. When making the determination, please provide the analysis process in the following format.
Note, for multiple-choice questions, it is only correct if the output includes all the corresponding options. (For example: If the reference answer is: A, C, and the model response is: C, the answer consistency is: No)
Note, if the model response does not specify a particular option, and its correctness cannot be determined, the answer consistency should be: No.
\end{minipage}
\\
 
\midrule
Example & 
\begin{minipage}[s][8.5cm]{1.5\columnwidth}
\textbf{Question:} 如图所示，AD是$\bigtriangleup ABC$的中线，点E在AC上，BE交AD于点F．当$\frac{AF}{FD}=\frac{1}{2}$时，$\frac{AE}{AC}= \underline{}$ \\
\textbf{Model Response:} 首先，我们需要找到中线AD，然后找到点E在AC上的位置。找到中线AD，可以通过勾股定理求解，$AD^2 = AD \times AC^2$。设$AD = x$, $AC = y$，则有 $x^2 = x \* y^2$。<STEP1>\\ 解这个方程，我们得到 $y = \sqrt{x}$。现在我们知道点E在AC上，所以E的横坐标为y，纵坐标为x。 \\ 接下来，我们需要找到点F在BD上的位置。由于F点在BD上，我们可以设F点在BD的一半，即$F = (x + y) / 2$。<STEP2>\\ 现在我们可以计算AF和FD的长度。$AF = x - F$, $ FD = FD - F$ 将已知的x和y代入公式，得到： $AF = x - (x + y) / 2 FD = (x + y) / 2 - (x + y) / 2$ 简化得到：$AF = x - x FD = x - x$ 这意味着$AF = FD$。<STEP3>\\ 现在我们可以求解题目中的条件：$AE / AC = 1 / 2$ 将已知的AF和FD代入公式，得到：$AE / AC = (x - x) / (\sqrt{x} \* y)$化简得到：$AE / AC = 1 / 2$所以，答案为$AE / AC = 1 / 2$ <STEP4>\\
\textbf{Reference Response:} $\frac{4\pi }{3}-2\sqrt{3}$ \\
\textbf{Judge Result:} The relevant paragraph in the Model Response that provides the answer is 
所以，答案为AE / AC = 1 / 2. \\
Understand the question in Chinese. The Chinese model response means:
Therefore, the answer is AE / AC = 1 / 2. \\
Next, let’s analyze whether the correct answer has been given.
The reference answer $\frac{1}{5}$ is a numerical answer, and the output given in the model response is 1/2. The consistency of the answers is: <No>.
\end{minipage} 
\\
\bottomrule
\end{tabular} 
\captionsetup{justification=centering}
\caption{Detail prompt for GradeGPT and it's instruction construction prompt by GPT4.}
\label{tab:promt_for_gradegpt}
\end{table*}


We also listed the prompts used by GradeGPT in Tables \ref{tab:promt_for_gradegpt}.

\subsection{Prompt For Cross-Lingual Prompting and En-CoT}
We have listed the specific prompts used for En-CoT and Cross-Lingual Prompt during actual execution in Table \ref{tab:cross_ling_prompt}.
Unlike the original Cross-Lingual Prompt paper, for experimental simplicity, we only adopted a single-turn format. However, this suffices to illustrate the varying inferential capabilities across different languages in current LMMs.

\section{CMMaTH Dataset Details}

\subsection{Data Collection Details}
To more clearly elucidate our data collection process, we have depicted the overall pipeline of data collection in Figure \ref{fig:data_collection}.

\subsection{Knowledge Point Details}
We provided detailed annotations of knowledge points for our dataset and conducted preliminary clustering of these knowledge points. The distribution of knowledge points in different clusters is as follows:
\begin{figure*}[t]
    \begin{center}
    \includegraphics[width=2.0\columnwidth]{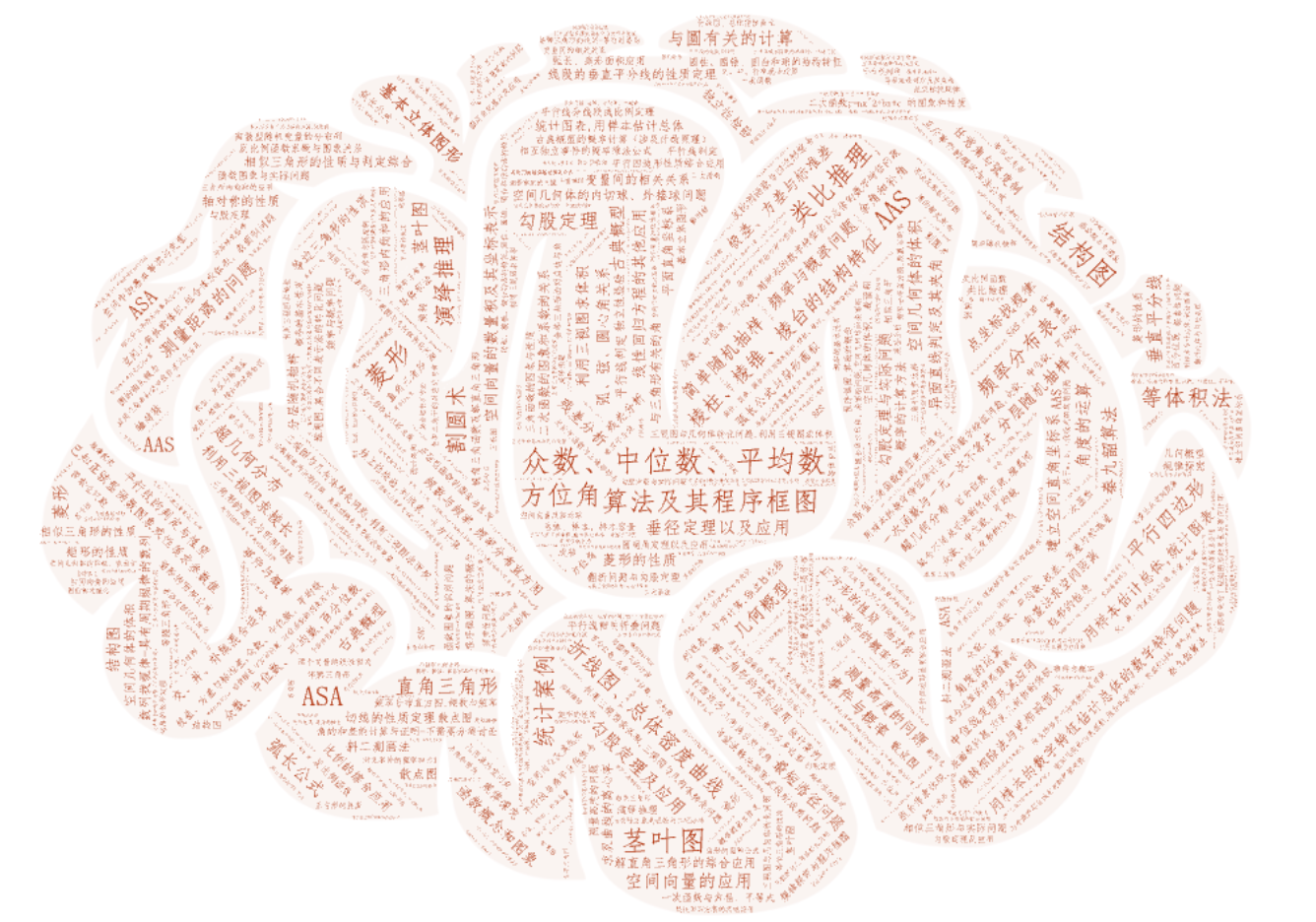} 
    \end{center}
    \caption{Cloud diagram of the knowledge points contained in the CMMaTH dataset.}
    \label{fig:wordbrain}
\end{figure*}
We have formulated a Knowledge \textbf{S}uccessful \textbf{S}olve \textbf{R}ate\textit{(SSR)} as a structural metric to gauge the proficiency level of multi-modal extensive models in mastering knowledge points. $N_{kn}$ is the total number of knowledge point of CMMaTH. $Acc_{kn_i}$ is the $Acc_{outcome}$ of questions about $i$'th knowledge point. $I$ denotes an indicator function.

\begin{equation}
    \small
    SSR@\alpha = \frac{\sum_{i=1}^{N_{kn}}I(Acc_{kn_i}>\alpha)}{N_{kn}} 
\end{equation}
It is our contention that a knowledge point can be deemed comprehensively understood only when the accuracy rate of solving problems related to that knowledge point surpasses a predefined threshold, denoted as $\alpha$. For the purpose of our investigation, we have established $\alpha$ at the values of 0.1, 0.2, 0.3, and 0.6 to demarcate the levels of mastery.

\subsection{Characteristics Of Annotators}
We utilized a standard team of four people, who spent two weeks annotating the data. All annotators have a university undergraduate education and are well-versed in basic knowledge of the K12 education field. To ensure quality, each question was verified by at least two people.

\section{GradeGPT details}
\label{GradeGPT_detail}
\subsection{GradeGPT Prompt Detail}
We have listed detailed Fewshot Examples using the GPT4-generated GradeGPT model responses in Table \ref{tab:cross_ling_prompt}. Through this table, you can observe the specific form of the Cross-Lingual-Judge-of-Chain that we have used.
\subsection{GradeGPT Performance Metric}

GradeGPT performance evaluation metric is precision in comparison. We constructed a model that responds to a test set containing outputs from various large models (including both correct and incorrect model outputs). Each output is labeled as correct or incorrect based on its result. GradeGPT is tasked with assessing whether the model responses are correct or incorrect, and this performance evaluation metric is a binary classification metric.
\begin{equation}
\small
    Acc_{outcome} = \frac{I(GradeGPT(R_i), Overcome_{GT})}{N_{response}}\times 100\
\label{eq: gradegpt_performace}
\end{equation}

\subsection{GradeGPT Training Details}
We generated cross-lingual evaluation instruction pairs using the outputs from InternLM-XComposer, LLaVA-v1.5, CogAgent-18B and Yi-VL-34B. These outputs were produced using GPT-4 Fewshot. The generated evaluation instructions were filtered based on specific rules, retaining only those responses from GPT-4 that contained the fields: <Yes>/<No>. Ultimately, we constructed a cross-lingual format instruction set comprising 56k instruction pairs.

\begin{figure*}[t]
    \begin{center}
    \includegraphics[width=2.0\columnwidth]{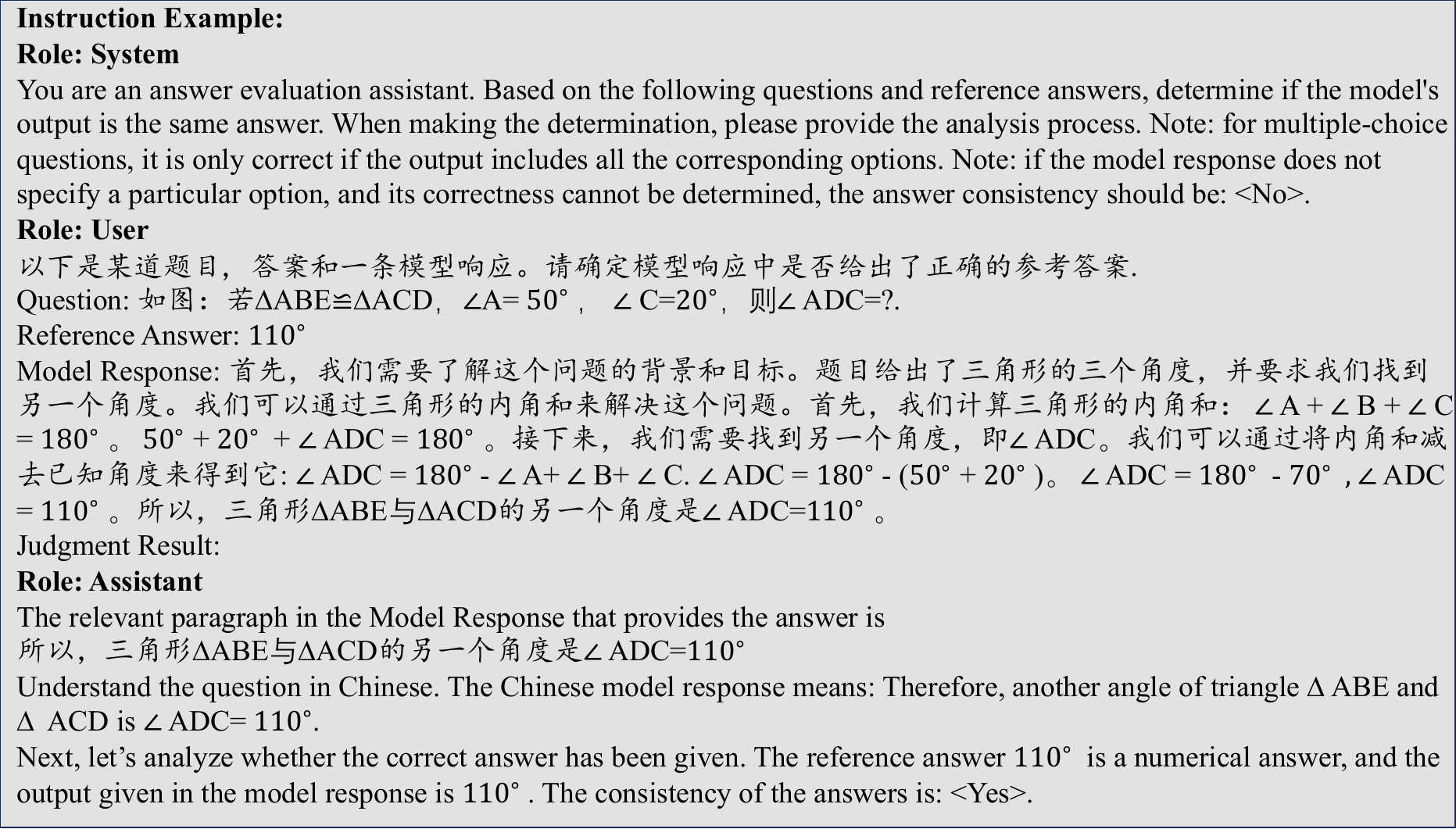} 
    \end{center}
    \caption{A instruction example to finetune GradeGPT.}
    \label{fig:instruction_example}
\end{figure*}

GradeGPT was trained on 8 H800, with the Qwen-14B-Chat version used as the base model. The model’s batch size was set to 16. The learning rate was set to 1e-4, and the gradient accumulation step was set to 16. It was trained for 10 epochs on a 40k bilingual Judge-of-Chain dataset. A detail example of instruction can refer to Figure \ref{fig:instruction_example}.

\begin{table}
    \small
    \centering
    \begin{tabular}{lcc}
        \toprule
        LLM   & $Acc_{outcome}$  \\
        \midrule
        Qwen-7B-Chat(4-Shot) &  35.1 \\
        \quad$+$Naive Outcome Finetune &  51.5 \\
        \quad$+$Judge-of-Chain & 65.3 \\
        \quad$+$Cross-Lingual-Judge-of-Chain &  85.1  \\   
        Qwen-14B-Chat(4-Shot) & 43.7 \\
        GradeGPT(14B)  &  96.1  \\
        \midrule
        GPT4(4-Shot) &  97.2  \\
        \bottomrule
    \end{tabular}
    \caption{Ablation study on the instruction fine-tuning of GradeGPT commands}
    \label{Aba:GradeGPT}
\end{table}

\subsection{Futher More Ablation Study}
We conducted experiments on a development set comprising outputs from a 0.5k model. The development set was sampled from a subset of 0.5k questions on CMMaTH. Each question was accompanied by answers provided by GPT-4V, GPT-4o, and middle school students. Each answer was manually annotated to indicate whether it was correct. We use \ref{eq: gradegpt_performace} to measure the answer judgment capability of different LMMs, including Zershot LMMs and LLMs after Finetune. \\
\textbf{Ablation On Instruction Format} We conducted experiments on various instruction enhancement techniques used by GradeGPT and compared the results with GPT4 in Table \ref{Aba:GradeGPT}. The results suggest that after various instruction enhancements, the accuracy of GradeGPT in model response judgment on CMMaTH can be improved to 96.1\%, significantly surpassing the accuracy of GPT4. The proposed strategy can significantly enhance GradeGPT’s ability to judge results. It is only slightly weaker than the performance of GPT4(Fewshot) executed with a large number of examples. Our GradeGPT, as an open-source parameter model of approximately 14B, can serve as a stable, low-cost, and efficient alternative to GPT4.

The Baseline we compared, Qwen-7B/14B(4-Shot), GPT4(4-Shot), \textit{Naive Outcome Finetune}, \textit{Judge-of-Chain}. In the \textit{Naive Outcome Finetune} format of instructions, the model is required to output its results indicating whether they are correct in the form of "<Yes>"/"<No>".. \textit{Judge-of-Chain} also includes the understanding of results and natural language descriptions of model outputs, but does not include the part of extracting key Chinese outputs and translating them into English. Compared to having the language model directly predict the <Yes>/<No> judgment labels, directly using Chinese Judge-of-Chain to construct Judge-of-Chain improves the model's performance in answer evaluation. However, they didn't yield good results. The performance disparity may stem from an imbalance in the quality of Chinese and English components within some bilingual base models due to the training corpus.

Our findings show that fine-tuning with Cross-Lingual-Judge-of-Chain for detailed thought chain refinement significantly improves the performance of open-source models in outcome analysis tasks. 
Additionally, we discovered that using bilingual thought chains instead of Chinese-only thought chains for base model fine-tuning effectively enhances performance in outcome determination tasks. By using and synthesizing the instructions in the form of Cross-Lingual-Judge-of-Chain that we designed, we are able to efficiently distill the answer reviewing capabilities of GPT4. \\
\textbf{Ablation On Instruction Data Source} 
The instruction data for Cross-Lingual Judge-of-Chain Prompts comes from outputs of various LLMs on CMMaTH. We conducted ablation experiments on the sources of instruction data, which showed the impact of using different LLM models on constructing diverse and effective instruction data.

\begin{table}
    \small
    \centering
    \begin{tabular}{lcc}
        \toprule
        Model Response Source   & $Acc_{outcome}$  \\
        \midrule
        LLaVA-v1.5 response &  77.2 \\
        \quad$+$InterLM-XComposer2-VL response &  83.1 \\
        \quad$+$CogAgent-Chat response & 84.2 \\
        \quad$+$Yi-VL-34B response & 85.1 \\
        \bottomrule
    \end{tabular}
    \caption{Ablation study on the instruction fine-tuning of GradeGPT commands. All experiments were conducted using the GradeGPT (7B) version.}
    \label{Aba:GradeGPT_source}
\end{table}

\section{Case Study}
\begin{figure*}[t]
    \begin{center}
    \includegraphics[width=2.0\columnwidth]{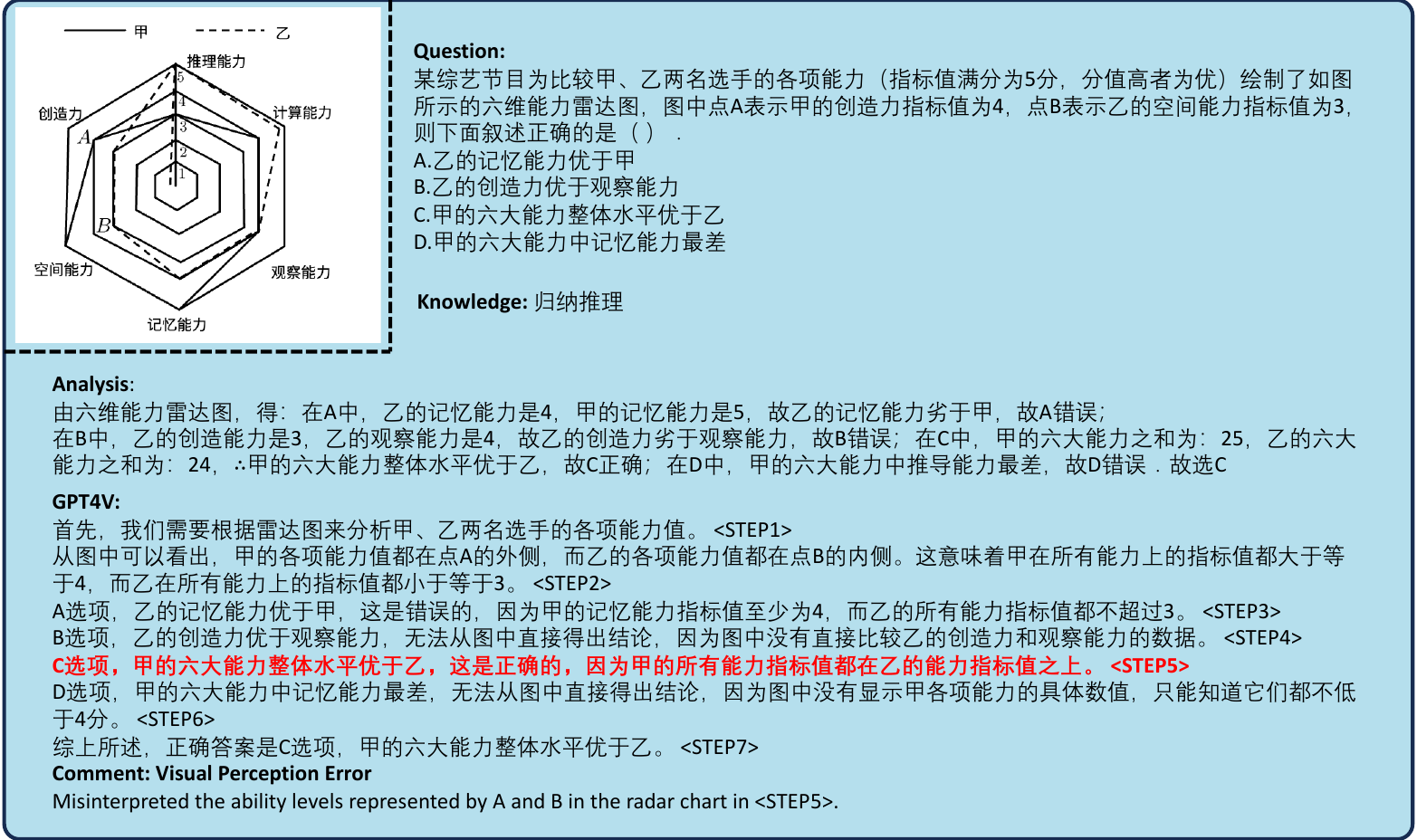} 
    \end{center}
    \caption{Case of GPT4V. The red ones are marked as generated inference hallucinations.}
    \label{fig:lidar_chart}
\end{figure*}

\begin{figure*}[t]
    \begin{center}
    \includegraphics[width=2.0\columnwidth]{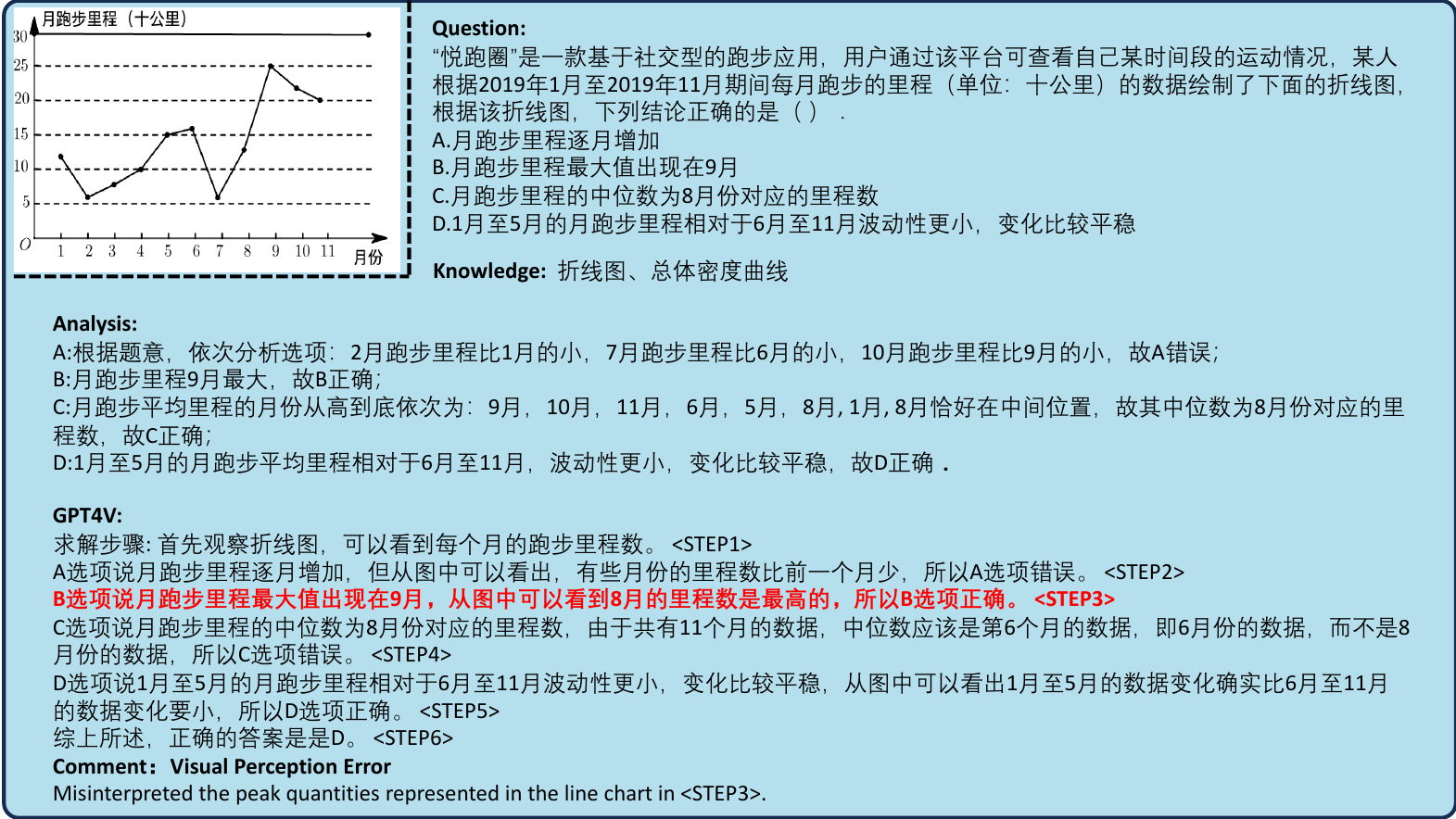} 
    \end{center}
    \caption{Case of GPT4V. The red ones are marked as generated inference hallucinations.}
    \label{fig:line_chart}
\end{figure*}

\begin{figure*}[t]
    \begin{center}
    \includegraphics[width=2.0\columnwidth]{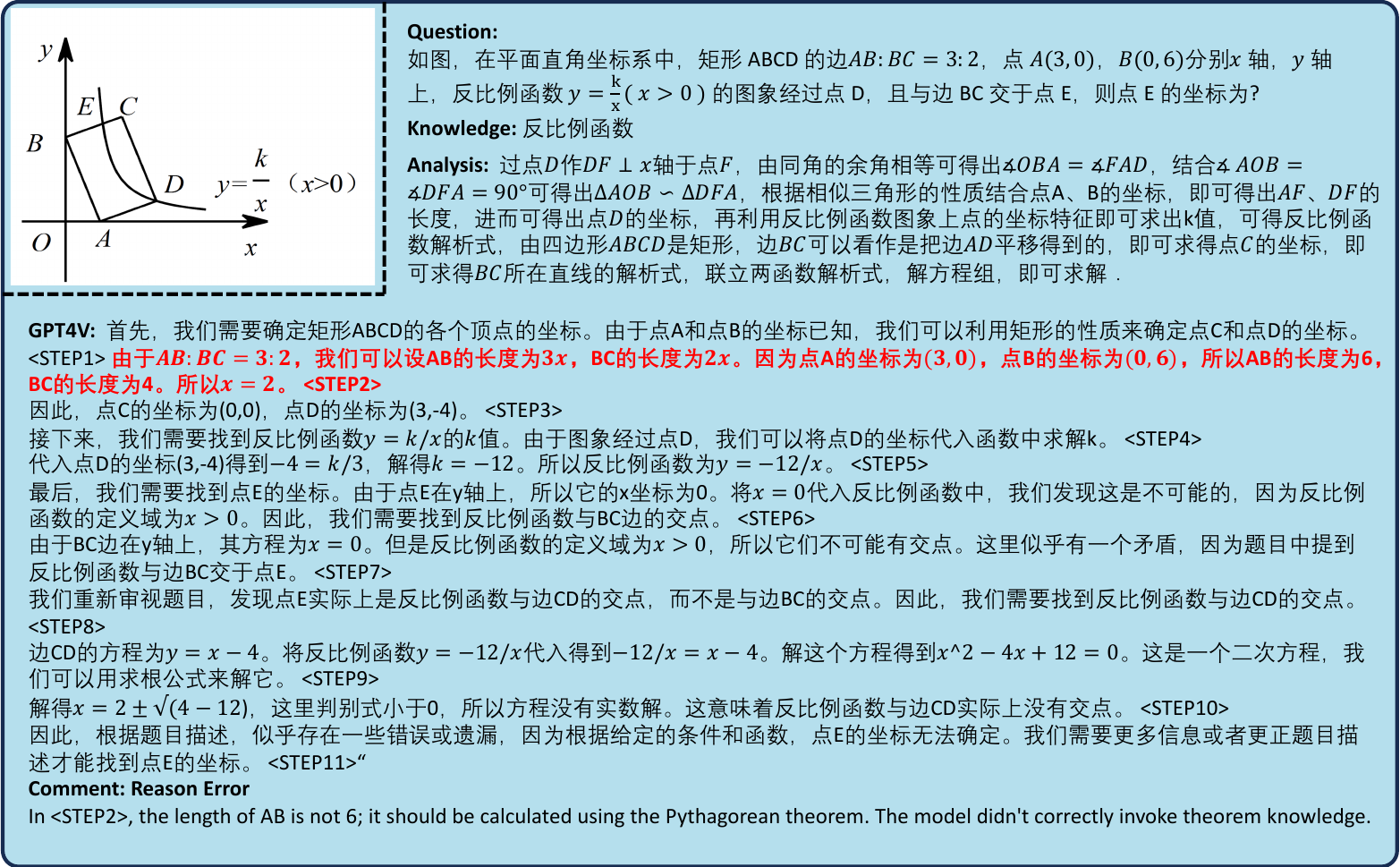} 
    \end{center}
    \caption{Case of GPT4V. The red ones are marked as generated inference hallucinations.}
    \label{fig:aba_geo_case_1}
\end{figure*}

\begin{figure*}[t]
    \begin{center}
    \includegraphics[width=2.0\columnwidth]{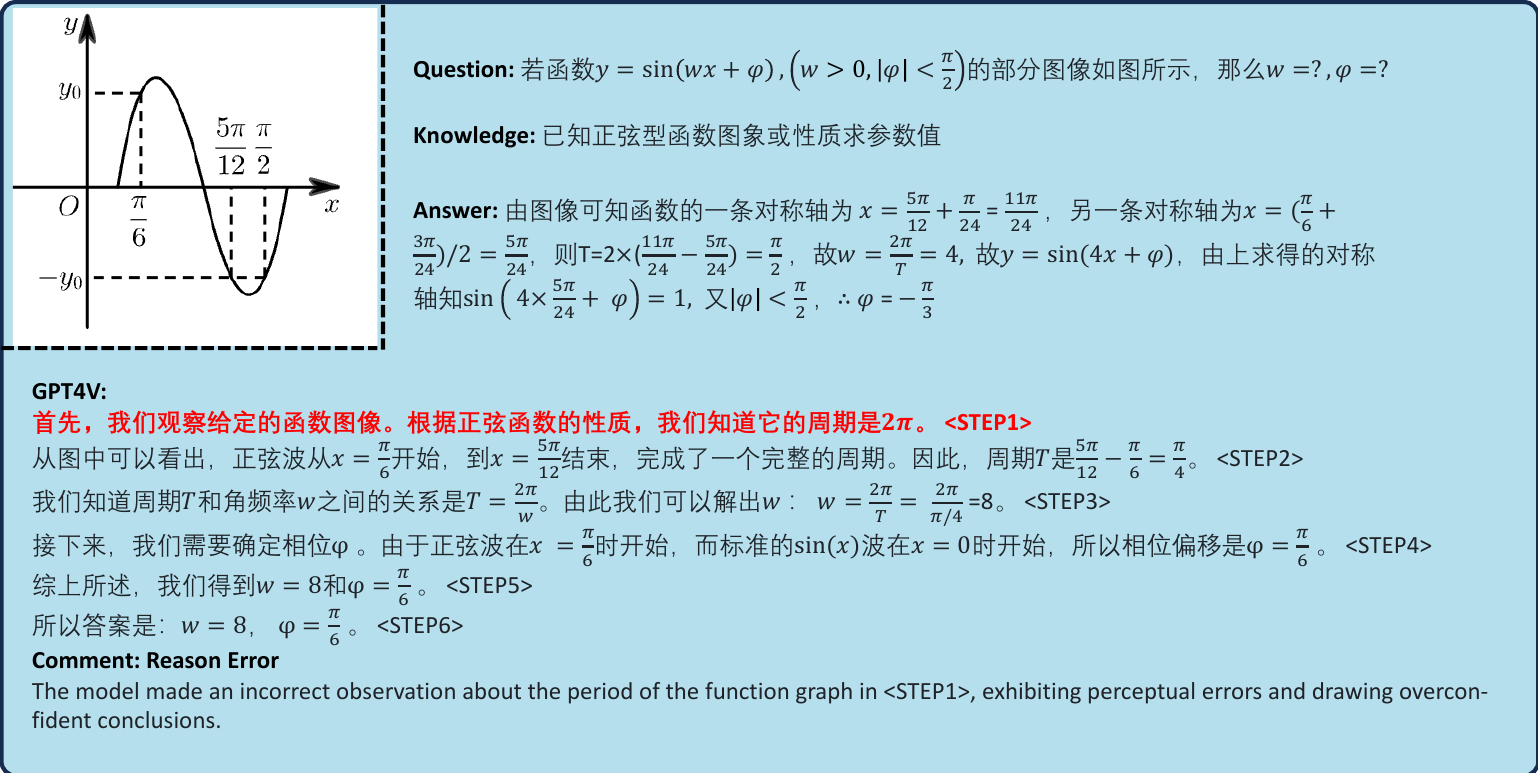} 
    \end{center}
    \caption{Case of GPT4V. The red ones are marked as generated inference hallucinations.}
    \label{fig:aba_geo_case_2}
\end{figure*}

\begin{figure*}[t]
    \begin{center}
    \includegraphics[width=2.0\columnwidth]{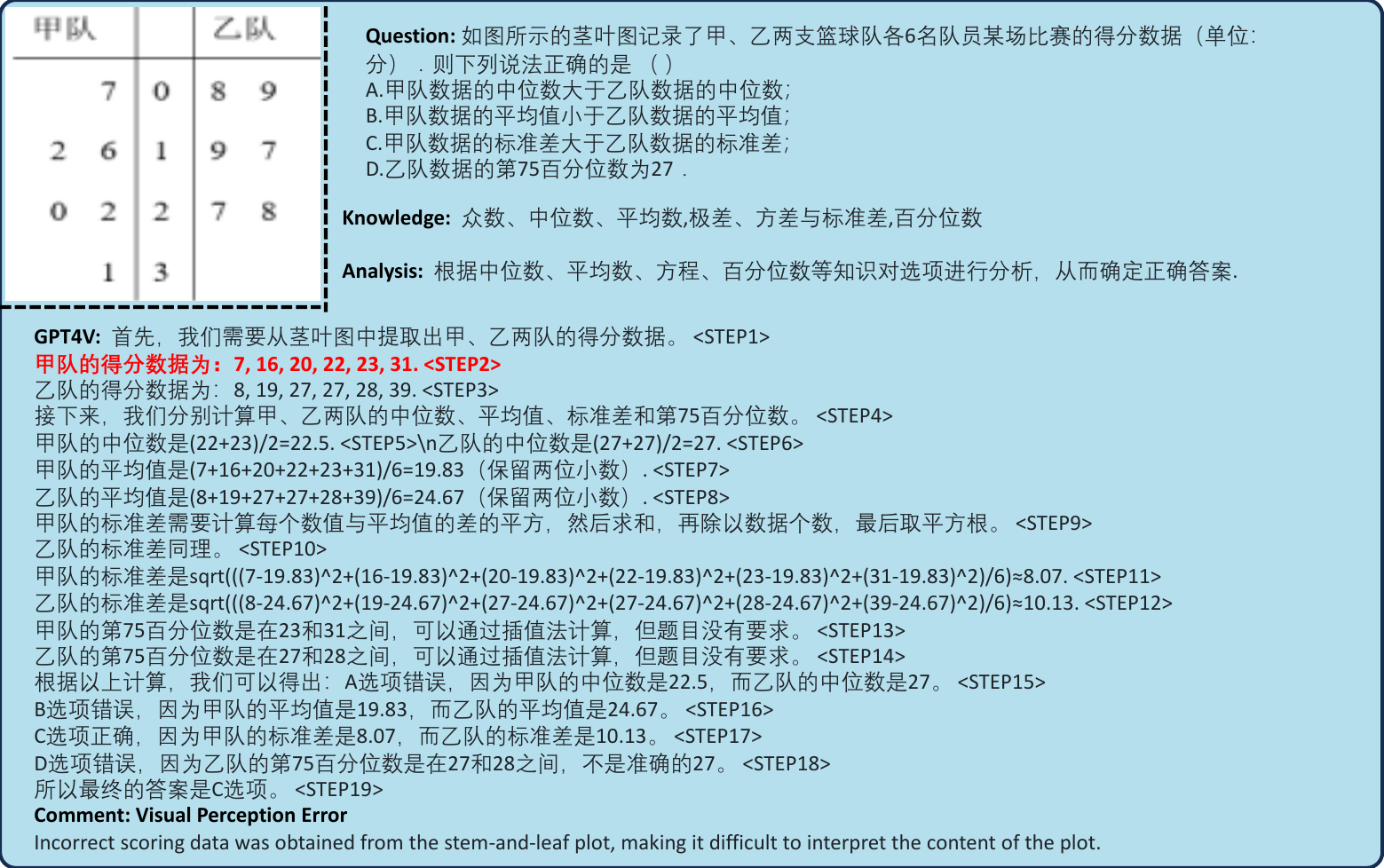} 
    \end{center}
    \caption{Case of GPT4V. The red ones are marked as generated inference hallucinations.}
    \label{fig:stem-and-leaf}
\end{figure*}

\begin{figure*}[t]
    \begin{center}
    \includegraphics[width=2.0\columnwidth]{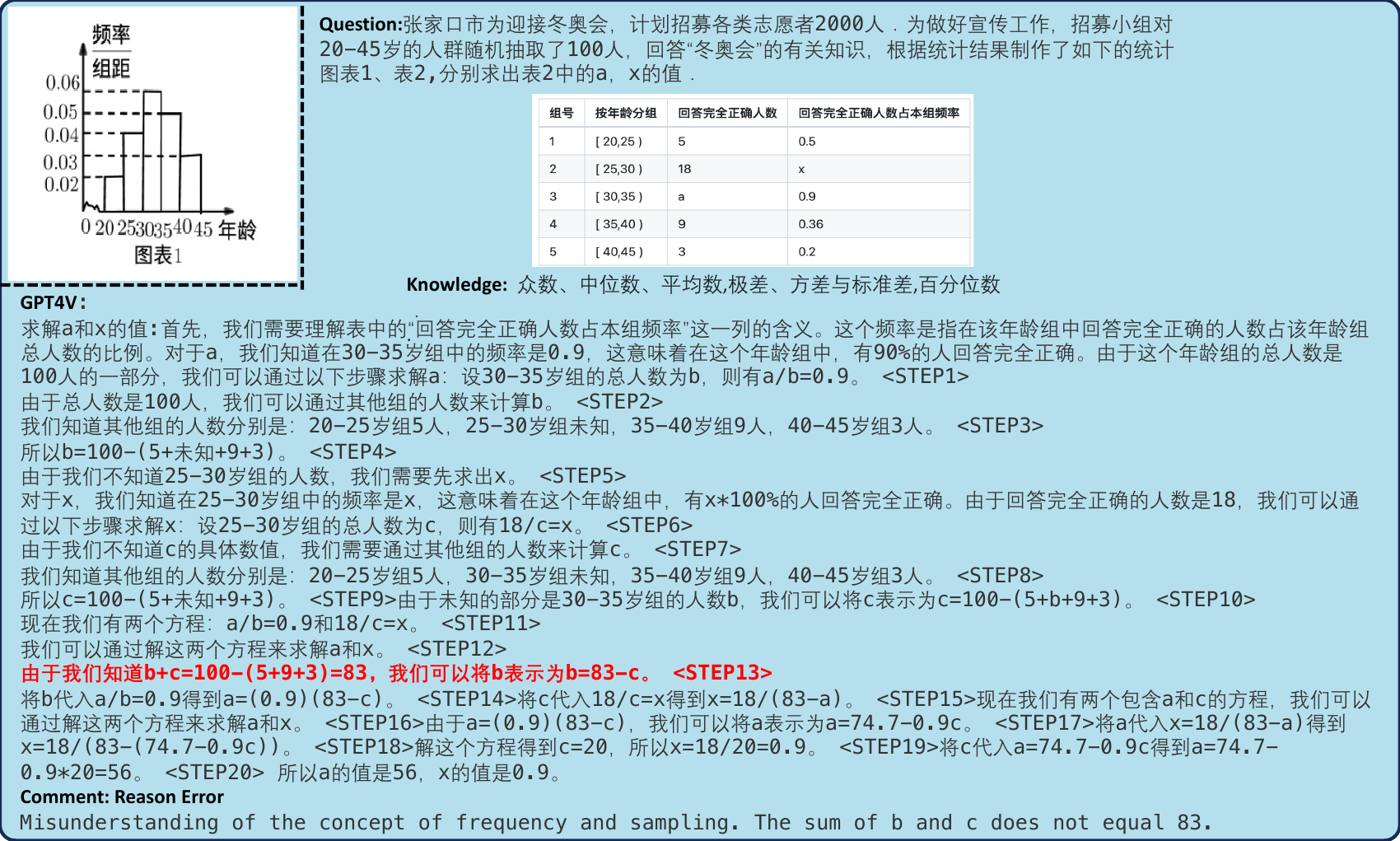} 
    \end{center}
    \caption{Case of GPT4V. The red ones are marked as generated inference hallucinations.}
    \label{fig:bar_chart}
\end{figure*}

\begin{figure*}[t]
    \begin{center}
    \includegraphics[width=2.0\columnwidth]{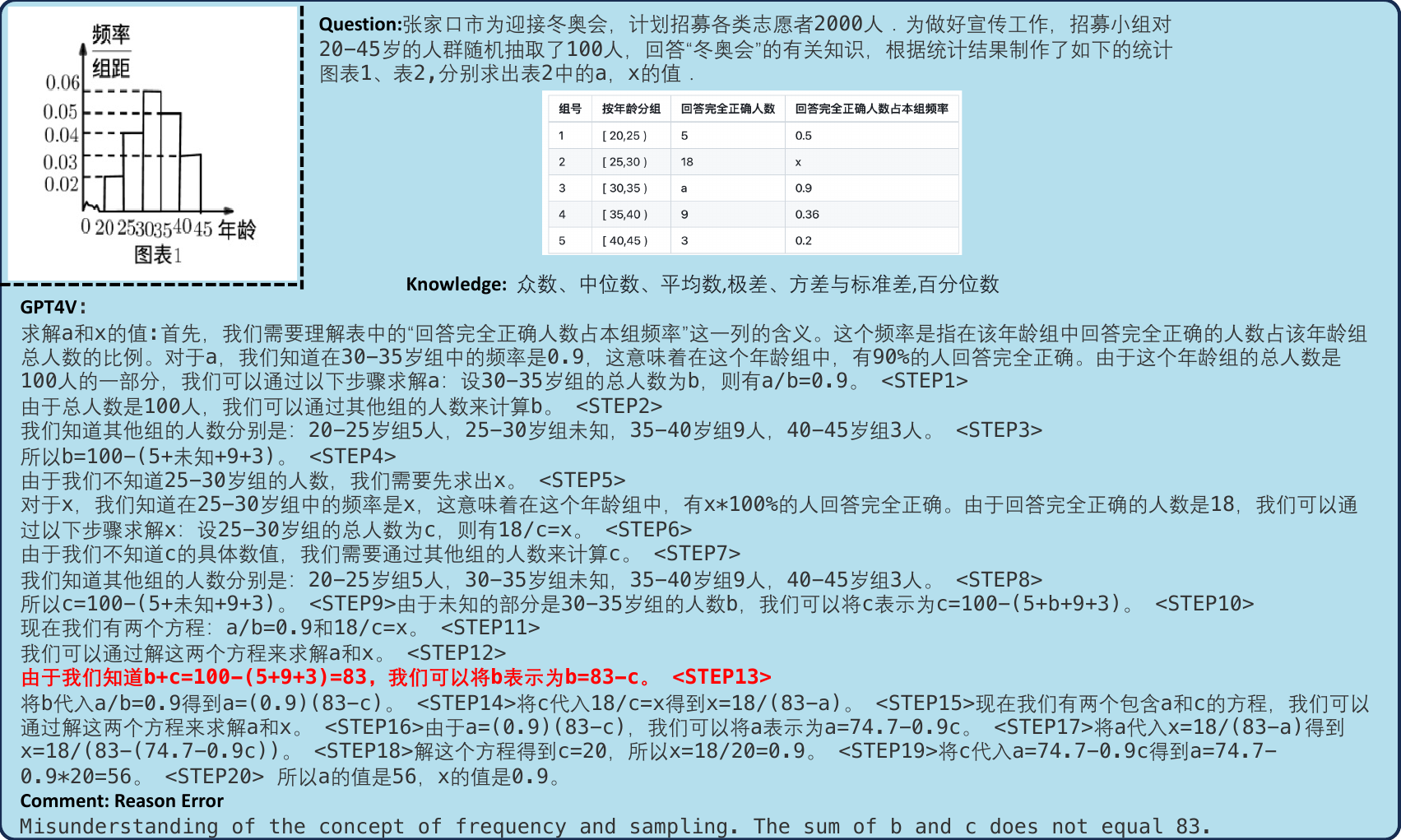} 
    \end{center}
    \caption{Case of GPT4V. The red ones are marked as generated inference hallucinations.}
    \label{fig:flow_chart}
\end{figure*}

\begin{figure*}[t]
    \begin{center}
    \includegraphics[width=2.0\columnwidth]{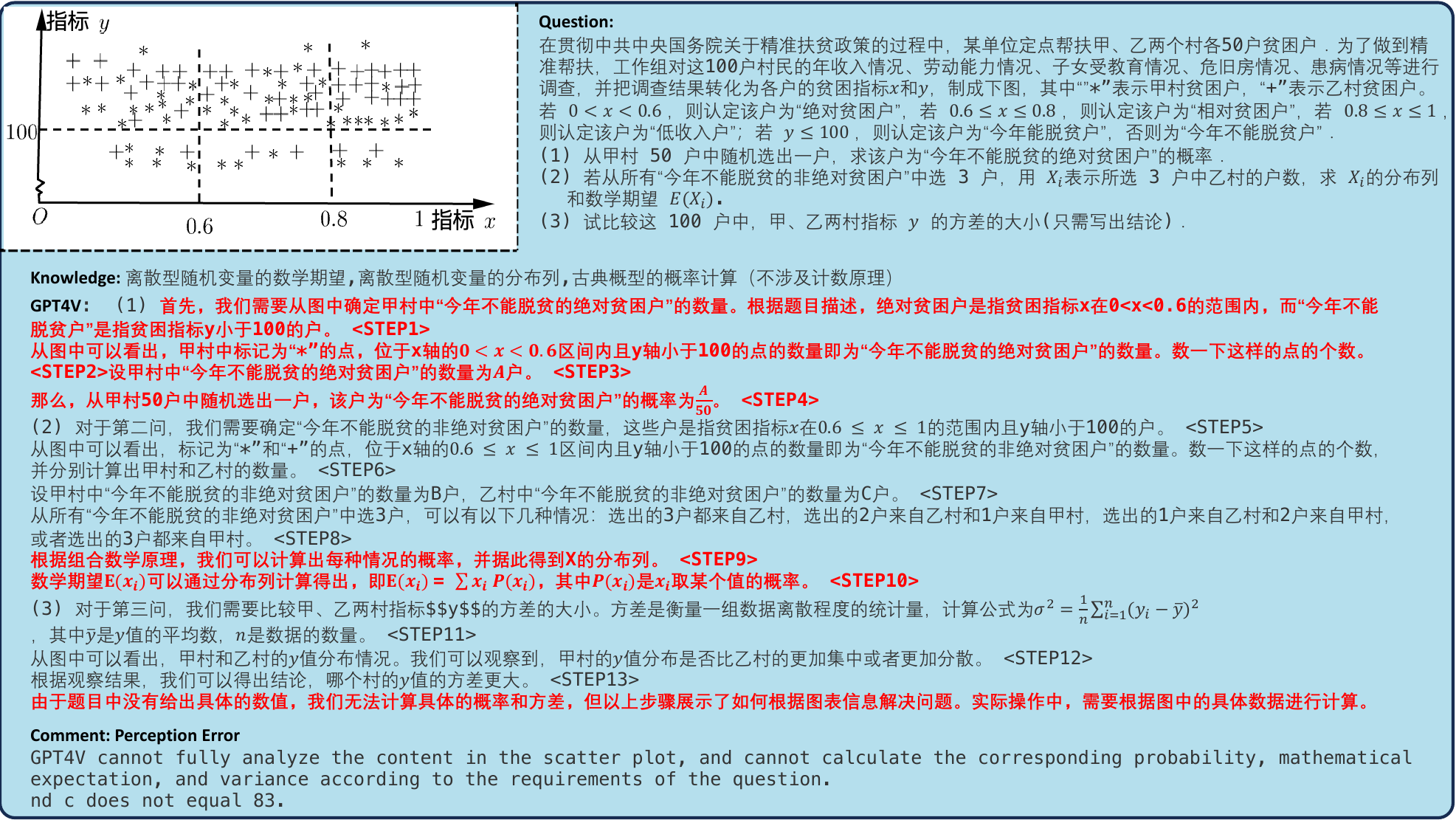} 
    \end{center}
    \caption{Case of GPT4V. The red ones are marked as generated inference hallucinations.}
    \label{fig:scatter_chart}
\end{figure*}

\begin{figure*}[t]
    \begin{center}
    \includegraphics[width=2.0\columnwidth]{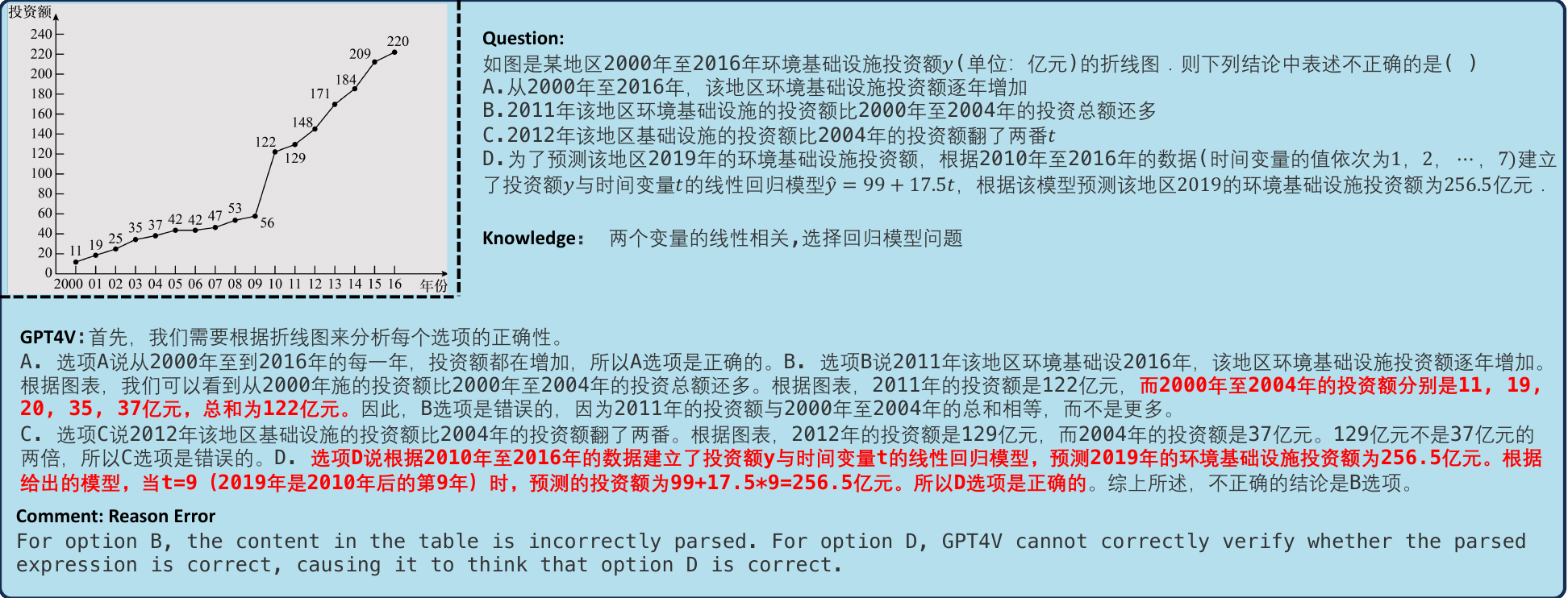} 
    \end{center}
    \caption{Case of GPT4V. The red ones are marked as generated inference hallucinations.}
    \label{fig:linear_chart}
\end{figure*}

\begin{figure*}[t]
    \begin{center}
    \includegraphics[width=1.3\columnwidth]{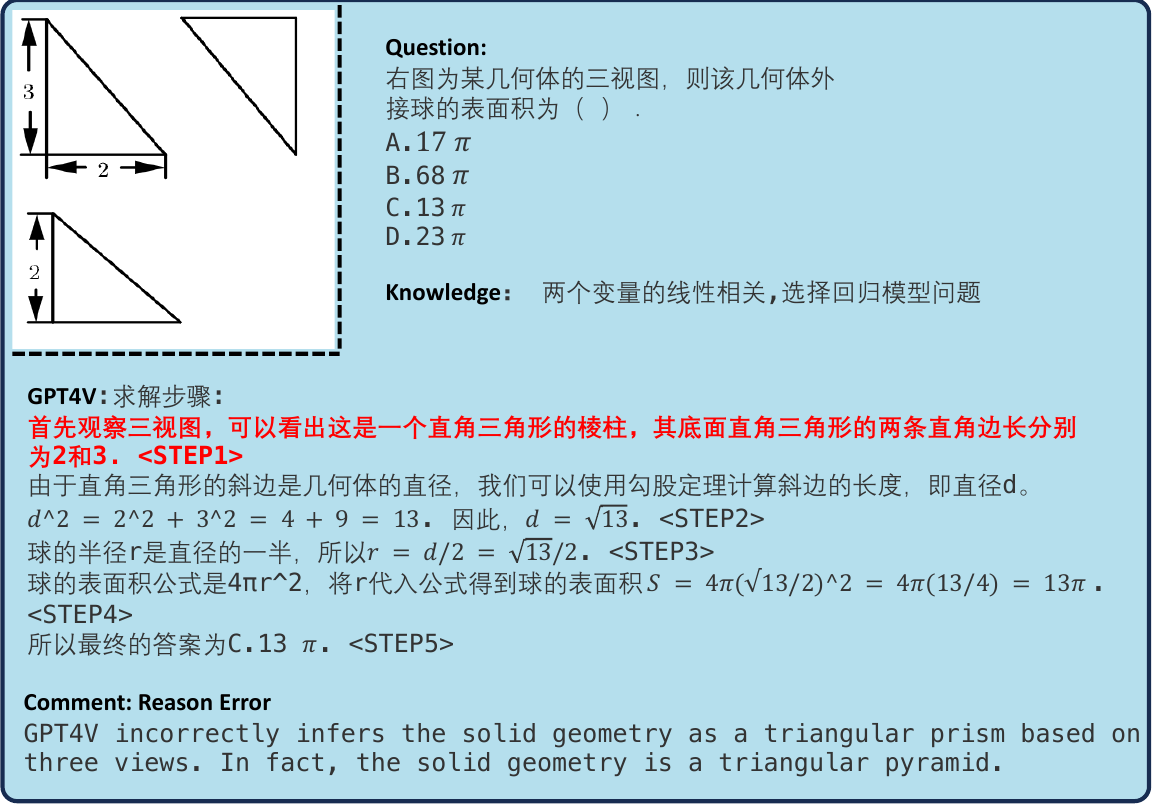} 
    \end{center}
    \caption{Case of GPT4V. The red ones are marked as generated inference hallucinations.}
    \label{fig:three_view_chart}
\end{figure*}

\begin{figure*}[t]
    \begin{center}
    \includegraphics[width=1.3\columnwidth]{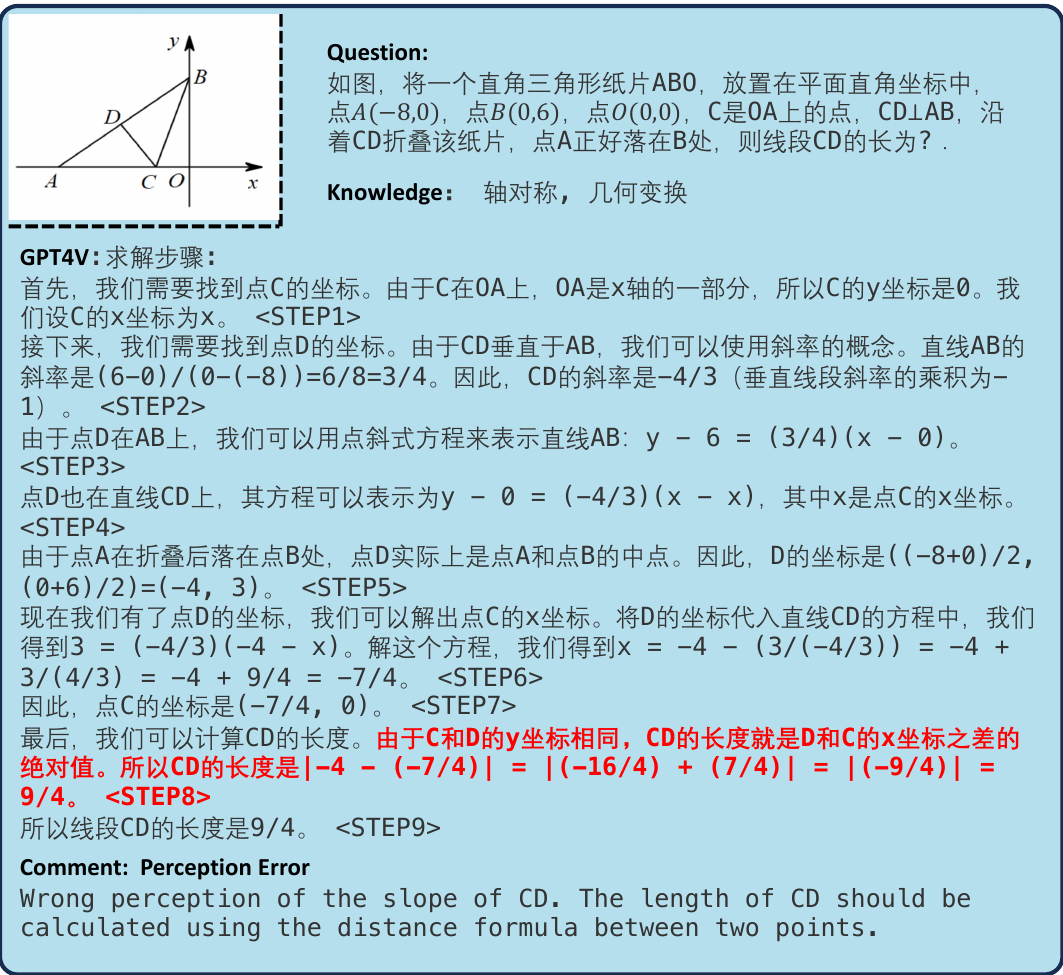} 
    \end{center}
    \caption{Case of GPT4V. The red ones are marked as generated inference hallucinations.}
    \label{fig:folded_image_chart_1}
\end{figure*}

\begin{figure*}[t]
    \begin{center}
    \includegraphics[width=1.0\columnwidth]{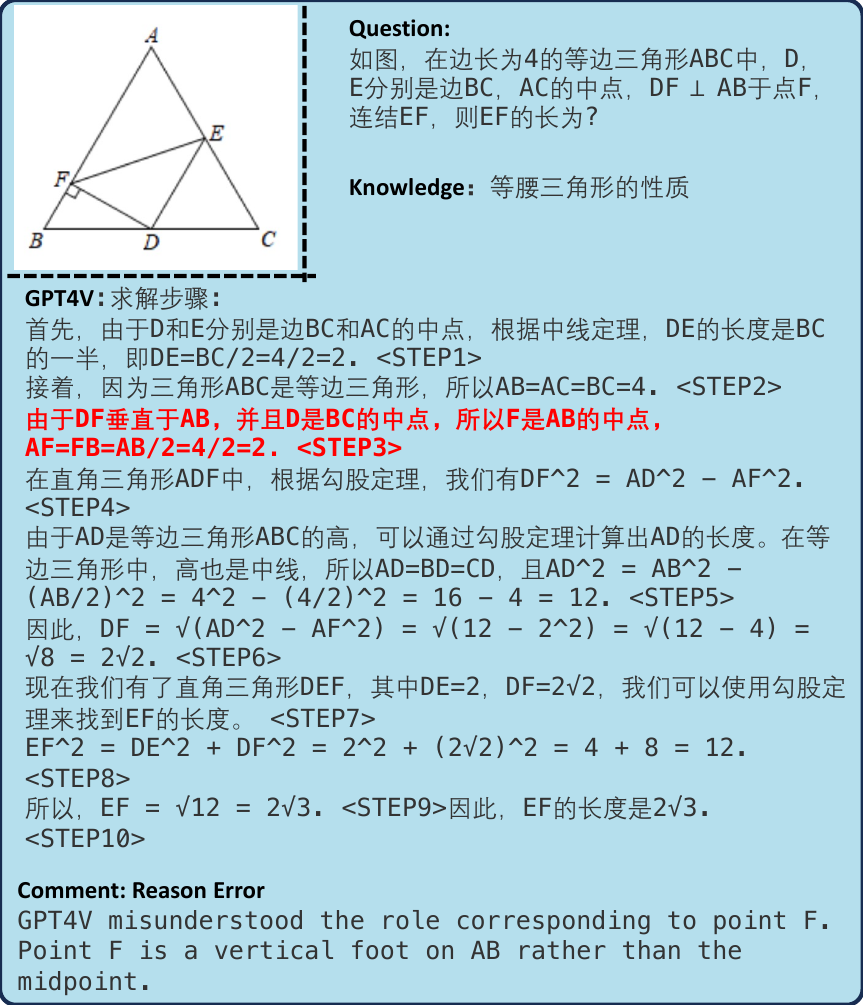} 
    \end{center}
    \caption{Case of GPT4V. The red ones are marked as generated inference hallucinations.}
    \label{fig:plane_geometry_chart}
\end{figure*}

\begin{figure*}[t]
    \begin{center}
    \includegraphics[width=1.3\columnwidth]{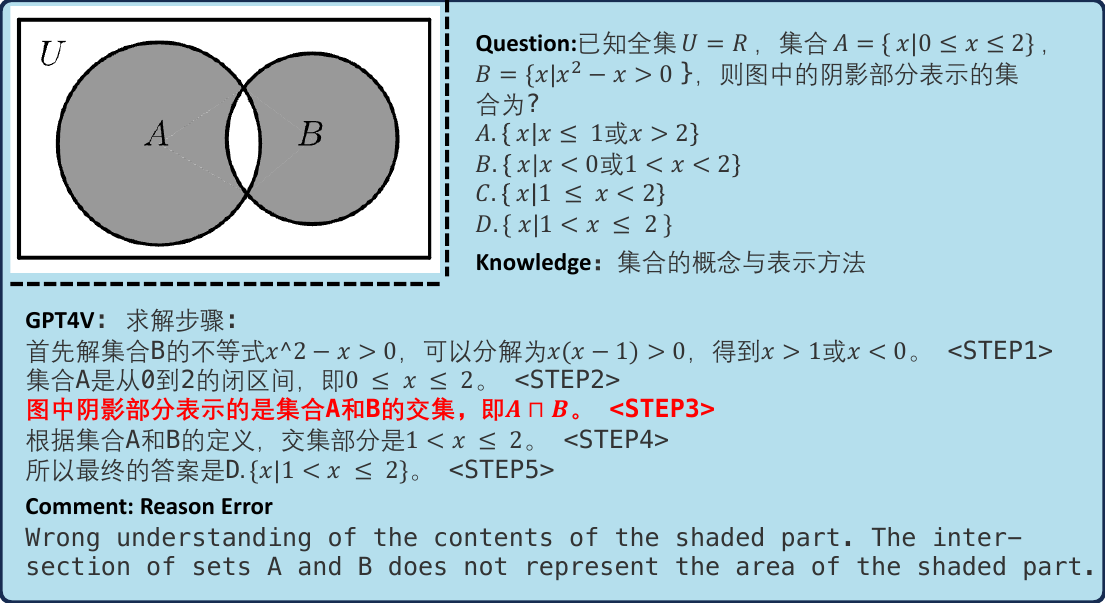} 
    \end{center}
    \caption{Case of GPT4V. The red ones are marked as generated inference hallucinations.}
    \label{fig:folded_image_chart_2}
\end{figure*}

We conducted a fine-grained manual evaluation of GPT4V's output on CMMaTH, with the results shown in Figure \ref{fig:lidar_chart} to Figure \ref{fig:folded_image_chart_2}.

\begin{table*}[tbhp!]
\centering
\small
\renewcommand\tabcolsep{2.5pt} 
\begin{tabular}{cp{13cm}}
\toprule
\textbf{Element} & \textbf{Prompt} \\
\midrule
system-prompt & 
\begin{minipage}[s][1.0cm]{1.5\columnwidth}
You are a math problem-solving assistant. Your input is a math problem and an image of the problem. Your task is to provide the solution steps and the final answer. The output format should be step-by-step.
\end{minipage}
\\
 
\midrule
Example & 
\begin{minipage}[s][1.8cm]{1.5\columnwidth}
\textbf{Question:} Xiaoming has 2000 yuan, an apple costs 2 yuan, Xiaoming bought 10 apples. How much money does Xiaoming have left? \\
\textbf{Model response:} First, calculate the cost of the apples Xiaoming bought, which is 2 times 10 equals 20 yuan. <STEP1>
Subtract the total cost from 2000 yuan, 2000-20=1980. <STEP2>
So the final answer is 1980 yuan. <STEP3>
\end{minipage}
\\
\bottomrule
\end{tabular} 
\captionsetup{justification=centering}
\caption{Prompt for En-CoT in Our Cross-Lingual Experiment.}

\label{tab:en_cot_prompt}
\end{table*}
\begin{table*}[tbhp!]
\centering
\small
\renewcommand\tabcolsep{2.5pt} 
\begin{tabular}{cp{13cm}}
\toprule
\textbf{Element} & \textbf{Prompt} \\
\midrule
system-prompt & 
\begin{minipage}[s][1.0cm]{1.5\columnwidth}
Please act as an expert in multi-lingual understanding in Chinese. After understanding, you should act as an expert in arithmetic reasoning
in English. Let's resolve the task you understand above step-by-step! Finally, you should format your answer as 'Answer: [num]'.
\end{minipage}
\\
 
\midrule
Example & 
\begin{minipage}[s][3.2cm]{1.5\columnwidth}
\textbf{Request:} 玛丽比琼小两岁，琼比杰莎大五岁。如果杰莎20岁，这三个女孩的年龄总和是多少？ \\
\textbf{Reason Result:} Let's understand the task in English step-by-step!
1. "玛丽比琼小两岁" means "Mary is two years younger than Joan."
2. "琼比杰莎大五岁" means "Joan is five years older than Jessie. "
3. "如果杰莎 20 岁" means "If Jessie is 20 years old. "
Then, Let's solve the task step-by-step:
1. According to the first sentence, Mary is two years younger than
Joan. Let's suppose Joan's age is J, then Mary's age is J - 2.
2. According to the second sentence, Joan is five years older than
Jessie. If Jessie's age is x, then Joan's age is x + 5.
Answer: 68.
\end{minipage}
\\
\bottomrule
\end{tabular} 
\captionsetup{justification=centering}
\caption{Detail prompt for Cross-Lingual Prompt(CLP) Reason technology for our Experiment.}

\label{tab:cross_ling_prompt}
\end{table*}

\begin{table*}[t]
\small
\centering
\begin{tabular}{cc}
\toprule
Model           & Name                             \\ \midrule
Llama-2-70B      & meta-llama/Llama-2-70b-hf       \\
MetaMath-70B     & meta-math/MetaMath-70B-V1.0     \\
DeepSeek-Math-7B & deepseek-ai/deepseek-math-7b-instruct  \\
Baichuan-13B     & baichuan-inc/Baichuan2-13B-Chat \\
Qwen-14B         & Qwen/Qwen-14B-Chat             \\
LLaVA-v1.5      & liuhaotian/llava-v1.5-13b       \\
InterLM-XComposer2-VL & internlm/internlm-7b       \\
Yi-VL-34B       & 01-ai/Yi-VL-34B       \\
CogAgent-Chat   & THUDM/cogagent-chat-hf            \\ \bottomrule
\end{tabular}
\caption{LLMs used in our experiments and their corresponding names in Huggingface Hub.}
\label{tab:model_huggingface_version}
\end{table*}

\begin{table*}[tbhp!]
\small
\centering
\begin{scriptsize}
\begin{tabular}{@{}lll@{}}
\toprule
\multicolumn{1}{c}{Model Name} & \multicolumn{1}{c}{Generation Parameters}                & \multicolumn{1}{c}{Comments}     \\ \midrule
Llama-2-70B                  & do\_sample=True, top\_k=0.5, top\_p=0.5, max\_tokens=512 & model=""Salesforce/codegen2-16B" \\ \midrule
GPT-4                          & temperature=0.2, max\_tokens=2048                         & version="gpt-4-1106-preview"     \\ \midrule
llava-7B-V1.5                  & temperature=0.2, max\_new\_tokens=2048                    & llava package                    \\ \midrule
DeepSeek-Math-7B                  & temperature=0.2, max\_new\_tokens=2048                    & vllm package                    \\ \midrule
Baichuan-13B                 & temperature=0.2, max\_new\_tokens=2048                    & vllm package                    \\ \midrule
Qwen-14B                 & temperature=0.2, max\_new\_tokens=2048                    & vllm package                    \\ \midrule
InterLM-XComposer2-VL        & temperature=0.2, max\_new\_tokens=2048                    &  Huggingface                    \\ \midrule
Yi-VL-34B        & temperature=0.2, max\_new\_tokens=2048                    & Huggingface                    \\ \midrule
CogAgent-Chat        & temperature=0.2, max\_new\_tokens=2048                    & Huggingface                    \\ \midrule
GPT4V                         & temperature=0.2, max\_tokens=2048                         & version="gpt-4-vision-2023-05-15"   \\ \bottomrule
GPT4o                         & temperature=0.2, max\_tokens=2048                         & version="gpt-4o-2024-02-01"   \\ \bottomrule
\end{tabular}
\caption{The hyperparameters for the models used in the evaluation are detailed. When the "comments" section includes the format \textit{model = ""}, it signifies that the model was loaded from the transformer package. The vLLM package indicates that models are implemented by the vLLM package, where more details can be found in \url{https://github.com/vllm-project/vllm}. For models other than OpenAI's GPT, custom codes were utilized for evaluation unless specified otherwise in the comments.}
\label{tab:model-hyperparameters}
\end{scriptsize}
\end{table*}

\end{CJK}
\end{document}